\documentclass[12pt]{article}
\usepackage{caption}
\usepackage{algorithm,algorithmicx,algpseudocode}
\usepackage{latexsym,amsmath,amssymb,amsfonts,amsthm,bbm}
\usepackage{natbib}
\usepackage{bibentry}
\usepackage[table]{xcolor} 
\usepackage[colorlinks=true,citecolor=blue]{hyperref}
\usepackage{enumerate}
\usepackage{subfig}
\usepackage{graphicx}
\usepackage{graphics,enumerate}
\usepackage{fancybox}
\usepackage{soul}
\usepackage{mathrsfs}
\usepackage{bm}
\usepackage{tikz}
\usetikzlibrary{shapes,snakes}
\usetikzlibrary{matrix}
\usepackage{multirow}
\usepackage{booktabs}
\usepackage{diagbox}
\newtheorem{theorem}{Theorem}

\newtheorem{lemma}{Lemma}

\newtheorem{assump}{Assumption}
\newtheorem{rmk}{Remark}

\let\counterwithin\relax
\usepackage{chngcntr}
\usepackage{apptools}
\AtAppendix{\counterwithin{lemma}{section}}

\newenvironment{eq*}{\begin{equation*}\begin{aligned}}{\end{aligned}\end{equation*}}

\def\supth{^{\scriptscriptstyle \rm th}}

\def\trans{^{\scriptscriptstyle \sf T}}

\def\diag {\mathop{\rm diag}}
\def\argmin{\mathop{\rm argmin}}

\def \eps {\varepsilon}
\def\hat{\widehat}

\makeatletter
\newcommand{\distas}[1]{\mathbin{\overset{#1}{\kern\z@\sim}}}%
\newsavebox{\mybox}\newsavebox{\mysim}
\newcommand{\distras}[1]{%
	\savebox{\mybox}{\hbox{\kern3pt$\scriptstyle#1$\kern3pt}}%
	\savebox{\mysim}{\hbox{$\sim$}}%
	\mathbin{\overset{#1}{\kern\z@\resizebox{\wd\mybox}{\ht\mysim}{$\sim$}}}%
}
\makeatother

\definecolor{red}{RGB}{200,50,150}
\definecolor{darkred}{RGB}{150,50,50}
\definecolor{brown}{RGB}{250,100,100}
\definecolor{green}{RGB}{000,150,100}
\definecolor{purple}{RGB}{250,000,180}

\def\red{\color{red}}

\def\red{\color{red}}

\newcommand{\bfm}[1]{\ensuremath{\mathbf{#1}}}

   \def\bA{\bfm A}  
   \def\bB{\bfm B}  
   \def\bC{\bfm C}  
   \def\bD{\bfm D}  
\def\be{\bfm e}   \def\bE{\bfm E}  \def\EE{\mathbb{E}}

\def\bh{\bfm h}   \def\bH{\bfm H}  
   \def\bI{\bfm I}  
     
   \def\bK{\bfm K}  
   \def\bL{\bfm L}  
   \def\bM{\bfm M}  
   \def\bN {\bfm N}  
   \def\bO{\bfm O}  \def\OO{\mathbb{O}}
   \def\bP{\bfm P}  
     
     \def\RR{\mathbb{R}}
   \def\bS{\bfm S}  
     
   \def\bU{\bfm U}  
\def\bv{\bfm v}   \def\bV{\bfm V}  
   \def\bW{\bfm W}  
   \def\bX{\bfm X}  
   \def\bY{\bfm Y}  
   \def\bZ{\bfm Z}  
                  \def\bN{\bfm N} 
  \def\bTheta{\bfm \Theta}

  \def\bOmega{\bfm \Omega}
\def\bSigma{\bfm \Sigma}
\def\bLambda{\bfm \Lambda}
\def\bDelta{\bfm \Delta}

\def\calD{{\cal  D}} 
 
\def\calF{{\cal  F}} 
\def\calG{{\cal  G}}

\def\calL{{\cal  L}}

 \def\scrO{{\mathscr  O}}
\def\calP{{\cal  P}} \def\scrP{{\mathscr  P}}
\def\calQ{{\cal  Q}} 
 
\def\calS{{\cal  S}} 
 
 \def\scrU{{\mathscr  U}}
\def\calV{{\cal  V}}

 \def\scrZ{{\mathscr Z}} 
\def\bzero{\bfm 0}
\def\bone{\bfm 1}

\def\nmin{n_{\scriptscriptstyle \sf min}}
\def\nmax{n_{\scriptscriptstyle \sf max}}

\def\hmax{h_{\scriptscriptstyle \sf max}}
\def\hmin{h_{\scriptscriptstyle \sf min}}
\def\thmin{\tilde{h}_{\scriptscriptstyle \sf min}}
\def\thmax{\tilde{h}_{\scriptscriptstyle \sf max}}

\def\MCE{{\sf MCE}}
\def \sfL{{\sf L}}

\def\bUhat{\widehat{\bU}}

\usepackage{footnotebackref}
\usepackage{setspace}

\begin{document}
		\title{Consensus Knowledge Graph Learning via Multi-view Sparse Low-rank Block Model}
		\author{Tianxi Cai$^{1,2*}$, Dong Xia$^{3*}$, Luwan Zhang$^{1*}$ and Doudou Zhou$^{4*}$ \\ \small
		$^1$Department of Biostatistics, Harvard T.H. Chan School of Public Health, Boston MA \\ \small
		$^2$Department of Biomedical Informatics, Harvard Medical School, Boston MA \\ \small
		$^3$Department of Mathematics, Hong Kong University of Science and Technology, Kowloon, Hong Kong \\  \small
  $^4$Department of Statistics and Data Science, National University of Singapore  \\ \small
		$^*$ alphabetical order}
		\date{}
		\maketitle
		
\begin{abstract}
Network analysis has been a powerful tool to unveil relationships and interactions among a large number of objects. Yet its effectiveness in accurately identifying important node-node interactions is challenged by the rapidly growing network size, with data being collected at an unprecedented granularity and scale. Common wisdom to overcome such high dimensionality is collapsing nodes into smaller groups and conducting connectivity analysis on the group level. Dividing efforts into two phases inevitably opens a gap in consistency and drives down efficiency. Consensus learning emerges as a new normal for common knowledge discovery with multiple data sources available. In this paper, we propose a unified multi-view sparse low-rank block model (msLBM) framework, which enables simultaneous grouping and connectivity analysis by combining multiple data sources. The msLBM framework efficiently represents overlapping information across large scale concepts and accommodates different types of heterogeneity across sources. Both features are desirable when analyzing high dimensional electronic health record (EHR) datasets from multiple health systems. An estimating procedure based on the alternating minimization algorithm is proposed. Our theoretical results demonstrate that a consensus knowledge graph can be more accurately learned by leveraging multi-source datasets, and statistically optimal rates can be achieved under mild conditions. Applications to the real world EHR data suggest that our proposed msLBM algorithm can more reliably reveal network structure among clinical concepts by effectively combining summary level EHR data from multiple health systems.

\end{abstract}
\vskip 10pt
{\bf Key words:} multi-view, consensus network, low-rank block model, sparse random effect

\newpage
\def\supth{^{\rm th}}
\def\subm{_{[m]}}
\section{Introduction}

Network analysis that unveils connectivity and interactions among a large number of objects is a problem of great importance with wide applications in social sciences, genomics, clinical medicine, and beyond \cite[e.g.]{goh2007human,nabieva2005whole,luscombe2004genomic,scott1988social}. As data are being collected at an unprecedented granularity and scale, it is now possible to study the structure of large networks. However, it is challenging to accurately infer the network structure in the presence of high dimensionality, especially when many nodes represent highly similar entities and multiple data sources are available.  A simple approach to overcome the high dimensionality and overlapping entities is to collapse similar nodes into groups. With groupings given as a priori, network connectivity analysis is subsequently performed on the group level to improve interpretability and reproducibility. For example, inferences for gene regulatory networks in genomics are often made on the pathway level that generally represents a group of functionally related genes \cite[e.g.]{kelley2005systematic, xia2018multiple}. Brain function network analyses are often performed on groups of voxels localized within a small region having a common neurological function \cite[e.g.]{shaw2007attention, chen2017shared, lu2017post}. However, for many applications, the group structure is unknown and needs to be estimated together with the network structure. In natural language processing (NLP),  synonymous terms should be grouped yet such grouping structure varies by context and is not generally available. In association studies linking current procedural terminology (CPT) codes to clinical outcomes, many procedures are clinically equivalent yet the currently available grouping of CPT codes is extremely crude \citep{ccs2019}.

Despite the potentially large sample size, network structure inferred from a single data source can be influenced by the characteristics or generation process of the data itself.
As more data sources become available, it is highly desirable to synthesize information from multiple sources, often termed as {\em views}, to jointly infer about a consensus network structure. The network structure achieved through consensus may differ from those derived from individual data sources, as each source might exhibit unique mechanisms or patterns specific to it, potentially leading to biases.  This debiasing for common knowledge discovery is particularly important when dealing with inherently heterogeneous data sources. A prime example comes from knowledge extraction using Electronic Health Records (EHR) data. The EHR system contains rich longitudinal phenotypic information from millions of patients. The EHR data is a valuable source for learning medical knowledge networks linking each specific disease condition with co-morbidities, diagnostic laboratory measurements, procedures, and treatment. Constructing a consensus network using data from multiple EHR systems could potentially remove bias due to different patient populations, physician training, and practices, as well as how or when the encodings are performed. However, the between-view heterogeneity also imposes methodological challenges to accurately learning network structure.  

To overcome these challenges, we propose in this paper a unified framework that can efficiently combine multi-view data to simultaneously group entities and infer about network structure.
We study methods for integrating several adjacency matrices from different views. Our underlying assumption is that after debiasing and standardizing each matrix to have unit diagonals, they converge toward a common latent correlation matrix $\bC$. The primary objective is to uncover a latent group structure within $\bC$. The complexity of this task arises from the variability in the unknown standardization weights and the noise levels present in the observed adjacency matrices, which can differ across views. We refer to this variability as `heterogeneity'. Our proposed estimation algorithm addresses this heterogeneity by identifying the signal-to-noise ratio specific to each view. This approach is distinct from conventional methods that merge information across views before determining the structure of $\bC$.
Specifically, we aim to learn a consensus network that reflects shared knowledge using a collection of $m$ independently-observed graphs on $n$ common vertex set $\calV=\{v_j\}_{j\in [n]}$,
$$
\calG\subm= \Big\{  \calG_s = \{\calV, \bW_s\}: \calV = \{v_j\}_{j\in [n]}, \bW_s = \big[W_s(j_1,j_2)\big]_{j_1\in[n]}^{j_2\in[n]} \Big \}_{s \in [m]} \,,$$ 
where $[n]=\{1,\cdots,n\}$, and  $W_s(j_1,j_2) \in \RR$ is the observed edge weight between node $v_{j_1}$ and node $v_{j_2}$ from the $s\supth$ view.
The vertex set $\calV$ admits a latent grouping structure shared across $m$ views in that there exists a unique non-overlapping $K$-partition:
$$
\calV = \cup_{k=1}^K \calV_k, \calV_k \cap \calV_l = \emptyset, \forall 1\leq k <l \leq K ,
$$ 
which can be equivalently represented by a 0/1 matrix
$\bZ= \big[Z(j,k)\big]_{j\in[n]}^{k\in[K]} , \mbox{where } Z(j,k) =  I(v_j \in \calV_k)$. To model the network structure while accommodating heterogeneity across views, we assume a flexible multi-view sparse low-rank model for $\{\bW_s\}_{s=1}^m$. This model utilizes a shared sparse consensus matrix $\bC$, which encapsulates the common network structure across all views. Simultaneously, it accommodates heterogeneity by varying the degrees of nodes to reflect distinct characteristics of each view and integrates sparse differences between views to capture the unique biases inherent to each one. The consensus matrix $\bC$ can be further decomposed into $\bC =  \bZ \bOmega \bZ^\top$, where $\bOmega$ is a {\it group-level sparse and low-rank weight matrix}. Our goal is to simultaneously learn $\bZ$ and $\bC$ from $\calG\subm$ in the presence of heterogeneity. 

The proposed framework is particularly appealing for knowledge graph modeling with multi-view data for several reasons. To illustrate this, consider our motivating example of knowledge extraction with multi-view EHR data where the nodes represent clinical concepts including disease conditions, signs/symptoms, diagnostic laboratory tests, procedures, and treatments. First, nodes within a group can effectively represent stochastically equivalent and interchangeable medical terms. For example, the clinical concepts ``coronary artery disease" and ``coronary heart disease" are used interchangeably by physicians but are mapped to two separate clinical concept unique identifies in the unified medical language system (UMLS) \citep{bodenreider2004unified}. Second, the structure of the group-level dependency captured by $\bOmega$ can be used to infer clinical knowledge about a disease. The consensus graph is particularly appealing as it removes biases from individual healthcare systems. Third, the improved estimation of the low-rank weight matrix $\bOmega$ and $\bZ$ through consensus learning also leads to a more accurate embedding representation for the nodes.

In a special case where $\bOmega$ is full-rank and each entry is non-negative and upper bounded by $1$, $\bC$ reduces to the well-known stochastic block model (SBM) \citep{holland1983stochastic} since $\bC$ characterizes the underlying Bernoulli distribution for entries in the observed adjacency matrix. More broadly, recovering $\bOmega$ shows a direct effort tapping into the network dynamics. For example, in clinical practices, complex diseases are often accompanied by a series of symptoms that may need multiple concurrent treatments. Therefore, learning a knowledge network of disease would greatly help support decision-making toward precision medicine.  Lastly, the decomposition on $\bC$ embodies efficient vector representations that enable groups of node embeddings. This provides a new embedding technique applicable in many areas, such as proteins, DNA sequences, and fMRI,  to expand their existing embedding family serving broader research needs \citep[e.g.]{asgari2015continuous, nguyen2016control,choi2016learning,ng2017dna2vec, vodrahalli2018mapping}.

With a single view, a simple approach to achieve this goal is to first perform grouping based on scalable clustering algorithms   \cite[e.g.]{shi2000normalized, ng2002spectral, newman2006modularity, bickel2009nonparametric, zhao2012consistency} 
 and then learn the network structure. 
However, dividing efforts into two phases inevitably opens a gap that could potentially create friction in consistency and drive down efficiency in transmitting information. For the sole purpose of recovering $\bZ$, SBM is perhaps one of the most developed frameworks that enjoys both straightforward interpretations and good statistical properties \citep[e.g.]{rohe2011spectral, lei2015consistency,nielsen2018multiple, abbe2017community, gao2017achieving}.  Central to SBM is the idea that the observed adjacency matrix is a noisy version of a rank-$K$ matrix with eigenvectors having exactly $K$ unique rows. Each unique row can be comprehended as a $K$-dimensional vector representation for nodes in that group. However, this assumption becomes too restrictive requiring the embedding dimension $r$ to be tied to the number of groups $K$. As $K$ grows (potentially with $n$), the embedding dimension desirably remains low. To this end, we generalize SBM by introducing a low-rank block model (LBM) on $\bC$ to allow $\bOmega$ to be low-rank, thus decoupling $r$ from $K$.  Extending to a multi-view setting, we further introduce a multi-view sparse low-rank block model (msLBM) to jointly model the faithfulness to the consensus and view-specific varying parts. One theoretical contribution in our proposed msLBM model is to perform a low-rank and sparse matrix decomposition with overlapping subspace on multiple noisy data sources. 

Recent years have witnessed a fast-growing literature on multi-layer network analysis \citep[e.g.]{levin2017central,le2018estimating,tang2017robust,
wang2019joint, jones2020multilayer,
paul2020spectral,lei2020consistent,jing2020community,arroyo2021inference,levin2022recovering}. For example, \cite{arroyo2021inference} considered multiple random dot product graphs sharing a common invariant subspace and \cite{wang2019common} decomposed the logistic-transformed multi-view expected adjacency matrices to a common part and individual low-rank matrices. \cite{levin2022recovering} proposed the weighted adjacency spectral embedding under the assumption that multi-layer networks share a common connectivity probability with a potential low-rank structure, while the noise distributions of different layers can be heterogeneous. Recently,  \cite{macdonald2022latent} proposed a latent space multiplex networks model in which part of the latent representation is shared across all layers while heterogeneity is allowed for the other part. 

Our msLBM model differs from those prior works in three crucial aspects. First, our msLBM allows node-wise heterogeneity on the consensus graph of each view/layer while the existing literature assumes SBM on each layer. Secondly, what is more important, our method allows view-wise heterogeneity on the consensus graph across different views/layers. This additional flexibility enables us to deal with heterogeneous data collected from different sources. Finally, our msLBM model introduces an additional sparse signal on each view/layer which is unexplainable by the low-rank consensus graph. Oftentimes, these sparse signals can capture uncommon network structures in each view/layer. These new ingredients in msLBM are motivated by the uniqueness of multi-view EHR data. 
Meanwhile, all these differences also make it more challenging to estimate the underlying consensus graph in our msLBM.

The rest of the paper is organized as follows. In Section \ref{sec:model}, we elaborate in more detail on the proposed low-rank block model and its extension accounting for heterogeneity arising in a multi-view setting. We then propose an alternating minimization-based approach in Section \ref{sec:method} to learn the consensus network that is easy and fast to implement in practice. Section \ref{sec:theory} provides all theoretical justifications. Simulations are given in Section \ref{sec:simulations} to demonstrate the efficacy and robustness of the proposed method. In Section \ref{sec:realdata}, we apply the proposed method to generate a new set of clinical concept embeddings and yield a very insightful Disease-Symptom-Treatment network on Coronary Artery Disease, by integrating information from a large digital repository of journal articles and three healthcare systems. Proofs on theories in Section \ref{sec:theory} are relegated to Appendix.

\section{Multi-view Sparse Low-rank Block Model }\label{sec:model}
		
\subsection{Notations}
Throughout, we use a boldfaced uppercase letter to denote a matrix and the same uppercase letter in normal font to represent its entries. We use a boldfaced lowercase letter to denote a vector and the same lowercase letter in normal font to represent its entries. Let $\bI_r$ denote the $r\times r$ identity matrix and $\bone_n$ denote the $n$-dimensional all-one vector. We let $\|\cdot \|_{\ell_2}$ denote vector $\ell_2$-norm. For any matrix $\bA$, let $\|\bA\|, \|\bA\|_{\rm F}$  denote its spectral norm and Frobenius norm respectively, $\|\bA\|_{\ell_1} = \sum_{i,j}|A_{i,j}|$, $\|\bA\|_{\ell_\infty} = \max_{i,j} |A_{i,j}|$,  $\lambda_j(\bA)$ denote its $j\supth$ largest singular value, $\bA_{i:}$ and $\bA_{:j}$ respectively denote its $i\supth$ row and $j\supth$ column, ${\rm Vec}(\bA)$ denote vectorizing $\bA$ column by column, $\kappa(\bA)= \lambda_1(\bA)/\lambda_{{\rm rank}(\bA)}(\bA)$ denote the condition number of $\bA$. For a set $\calV$, we use ${\rm Card}(\calV)$ to denote its cardinality. 
We denote the set of $r\times r$ orthonormal matrices by
$$
\scrO_{r, r}:=\big\{\bO\in\RR^{r\times r}: \bO \bO^{\top} = \bO^{\top}\bO=\bI_r\big\} .
$$
If the entries of an orthonormal matrix $\bO\in\scrO_{r,r}$ are either $0$ or $1$ such that each row and column contains one single nonzero entry, then we call $\bO$ a {\it permutation matrix}. The set of all $r\times r$ permutation matrices is denoted by $\scrP_{r,r} \subset \scrO_{r, r}$. 
With slight abuse of notations, we denote the set of $n\times K$ matrices with orthonormal columns by 
$$
\scrO_{n, K}:=\big\{\bX\in\RR^{n\times K}: \bX^{\top}\bX=\bI_K \big\} \,.
$$
Given any matrix $\bA\in\RR^{n\times r}$ with ${\rm rank}(\bA)=r$, let $\calP_{\bA}$ denote the orthogonal projection from $\RR^{n}$ to the column space of $\bA$, or more specifically $\calP_{\bA}(\bv)=\bA(\bA^{\top}\bA)^{-1}\bA^{\top}\bv$ for any $\bv\in\RR^{n}$. We denote $\be_{j}$ the $j$-th canonical basis vector, whose dimension might change at different appearances. Denote $\bA\circ \bB$ the Hadamard product of matrix $\bA$ and $\bB$, i.e., $(\bA\circ \bB)_{ij}=\bA_{ij}\bB_{ij}$. 

\subsection{Low-rank Block Model (LBM)}\label{sec:LBM}
We first introduce the graph model for a single view. 
Let $\calG=\{\calV, \bW \}$ denote an undirected weighted graph with vertex set $\calV=\{v_j\}_{j\in [n]}$ and symmetric weight matrix $\bW = [W(j_1,j_2)]_{j_1 \in [n]}^{j_2 \in [n]}$ with $W(j_1,j_2) \in \RR$ representing the connection intensity between the vertices $v_{j_1}$ and $v_{j_2}$. We assume that the graph $\calG$ admits a latent network structure in the sense that there exists a unique (unknown) non-overlapping $K$-partition of the vertex set $\calV$:
$$
\calV = \cup_{k=1}^K \calV_k, \quad {\rm Card}(\calV_k)=n_k, \quad \calV_k \cap \calV_l = \emptyset, \quad \forall 1\leq k <l \leq K, 
$$
with $n=\sum_{k=1}^K n_k$, which can be equivalently represented by a {\it group membership matrix} 
$$
\bZ= \big[Z(j,k)\big]_{j\in[n]}^{k\in[K]} , \quad \mbox{where}\quad Z(j,k) =  {\bf 1}(v_j \in \calV_k)\in\{0,1\}.
$$
We denote by  $\scrZ_{n,K}$ the set of all possible $n\times K$ dimensional $K$-group membership matrices for $n$ nodes. Throughout, we assume $K$ is known that can grow with $n$ for high-dimensional cases for all theoretical analyses. Strategies for choosing an appropriate $K$ would be discussed in Section \ref{sec:realdata}. The set of orthornormalized membership matrices is denoted by
$$\scrU_{n, K} = \Big \{\bU := \bZ \left[\diag(\bone_n^{\top} \bZ)\right]^{-1/2}: \bZ \in \scrZ_{n,K} \Big\} \subset \scrO_{n,K} .$$
We assume the observed edge weight matrix $\bW$ can be decomposed as
\begin{equation}
	\label{eq: W decomp}
	 \bW = \bH \bZ \bOmega \bZ^\top \bH + \bE : = \bH \bC \bH	+ \bE  : = \bL	+ \bE ,
\end{equation}
where the symmetric matrix $\bOmega \in \RR^{K\times K}$ is the {\it group-level correlation matrix} that measures the strength of connectivity between groups with diagonal entries being $1$ and other entries bounded by $1$, the diagonal matrix $\bH={\rm diag}(h_1,\cdots,h_n)$ is the degree parameter that  
indexes the information contained by each node, and $\bE$ represents the sampling error. In a special case where each entry of $\bOmega$ is non-negative, and $\bH$'s entries are upper bounded by $1$, model (\ref{eq: W decomp}) reduces to the degree corrected stochastic block model (DCBM) \citep{karrer2011stochastic}  under which $W(j_1, j_2)$ represents the probability of $v_{j_1}$ and $v_{j_2}$ being connected, and the matrix $\bOmega$ is assumed to be full-rank to recover $\bZ$ for community detection. However, this full-rank assumption is inappropriate for knowledge graph modeling where $K$ is often large but $\bOmega$ is low rank. We instead assume the following a low-rank block model (LBM)  as a generalization of DCBM.
\begin{assump}[LBM] \label{assump:lowrank}
	The graph $\calG=(\calV,\bW)$ satisfies (\ref{eq: W decomp}) with $r:={\rm rank}(\bOmega) {\red \leq} K < n$ and $\bOmega$ is a positive semi-definite matrix.
\end{assump}

\begin{rmk}
The low-rank assumption on $\bOmega$ has been previously explored in latent space models, as reviewed by \cite{athreya2018statistical}. These studies allow for distinguishing between the dimensions of embeddings and the number of clusters within a network. Notably, \cite{tang2022asymptotically} investigated the effectiveness of spectral estimators within this framework. Our model, however, diverges from these earlier approaches by not relying on the independent edge assumption \citep{athreya2018statistical} and by accommodating weighted edges. 
Drawing inspiration from the application of learning clinical knowledge graphs, as detailed in Section \ref{sec:discuss}, we adopt the assumption that the group correlation matrix $\bOmega$ is positive semi-definite and symmetric. As further discussed in Section \ref{sec:discuss}, this assumption is imposed simply for presentation clearness. Our methods can be easily adapted for asymmetric matrix $\bOmega$, and similar theoretical results continue to hold.
\end{rmk}

Due to the additional heterogeneity $\bH$, the correlation matrix $\bC$ and the expected weight matrix $\bL$ can have drastically different eigenstructures. By definition and Assumption~\ref{assump:lowrank}, there exists a matrix $\bU\in\RR^{n\times r}$ such that $\bC=\bU\bU^{\top}$. Since the diagonal entries of $\bC$ are all ones, we have $\|\be_j^{\top}\bU\|=1$ for all $j\in[n]$. The row-wise separability of $\bU$ is immediately guaranteed by Lemma~\ref{lem:Urow_2}. We note that the columns of $\bU$ are not orthonormal. 
\begin{lemma}\label{lem:Urow_2}
Under Assumption~\ref{assump:lowrank}, for any $1\leq k_1<k_2\leq K$ and $j_1\in\calV_{k_1}$ and $j_2\in\calV_{k_2}$, we have 
	$$
	\|U(j_1,:)-U(j_2,:)\|_{\ell_2}\geq \delta_{\bOmega}\cdot \frac{\min_{k\in[K]} n_k^{1/2}}{\lambda_1^{1/2}(\bC)},
	$$
where $\delta_{\bOmega}:=\min_{1\leq k_1<k_2\leq K} \big\|\Omega(k_1,:)-\Omega(k_2,:) \big\|_{\ell_2} >0$.
\end{lemma}
\begin{rmk}
The condition $\delta_{\bOmega}>0$ is necessary since otherwise there exist $k_1,k_2\in[K]$ such that $\Omega(k_1,:)=\Omega(k_2,:)$ implying that there are no differences between the $k_1$-th group and $k_2$-th group. In that case, it is more reasonable to merge these two groups. 
\end{rmk}

We now study the eigenvectors of the matrix $\bL$. Recall that 
$$
\bL = \bH\bZ\bOmega\bZ^{\top}\bH=\big(\bH\bZ\bN_{\bH}^{-1}\big)\cdot (\bN_{\bH}\bOmega\bN_{\bH}) \cdot (\bH\bZ\bN_{\bH}^{-1})^{\top}
$$
where the diagonal matrix $\bN_{\bH}$ is defined by 
\begin{equation}\label{eq:tilde_hk}
\bN_{\bH}={\rm diag}(\tilde h_1, \tilde h_2,\cdots,\tilde h_K)\quad {\rm where}\quad \tilde h_k^2=\sum\nolimits_{j\in \calV_k}h_j^2 \quad {\rm for}\ k=1,\cdots,K.
\end{equation}
By definition, the matrix $\bH\bZ\bN_{\bH}^{-1}$ has orthonormal columns in that $(\bH\bZ\bN_{\bH}^{-1})^{\top}(\bH\bZ\bN_{\bH}^{-1})=\bI_{K}$.  Now, under Assumption~\ref{assump:lowrank}, consider the eigendecomposition of $\bN_{\bH}\bOmega\bN_{\bH}=\bV \bD \bV^{\top}$ 
with the eigenvectors $\bV \in\RR^{K\times r}$ having orthonormal columns corresponding to the eigenvalues $\bD={\rm diag}(d_1,\cdots,d_r)$ where $d_1 = \lambda_1(\bD)\geq \cdots \geq d_r = \lambda_r(\bD)>0$. Then we may obtain the eigendecomposition of $\bL$ as
\begin{equation}\label{eq:UDU}
\bL = \bar{\bU} \bD \bar{\bU}^{\top}, \quad \mbox{where } \bar{\bU} =\bH \bZ\bN_{\bH}^{-1}\bV.
\end{equation}
The row-wise separability of $\bar\bU$ is given by Lemma~\ref{lem:Urow}. 
\begin{lemma}\label{lem:Urow}
	Under Assumption~\ref{assump:lowrank}, for any $1\leq k_1<k_2\leq K$ and $j_1\in\calV_{k_1}$ and $j_2\in\calV_{k_2}$, we have 
	$$
	\|\bar{U}(j_1,:)-\bar{U}(j_2,:)\|_{\ell_2}\geq \frac{\delta_{\bH,\bOmega}}{\lambda_1(\bOmega)}\cdot \frac{\thmin}{\thmax^2}
	$$
	where $\delta_{\bH,\bOmega}:=\min_{1\leq j_1<j_2\leq n} \big\|(\be_{j_1}-\be_{j_2})^{\top}\bH\bZ\bOmega \big\|_{\ell_2}$, $\thmin=\min_k \tilde{h}_k$, and $\thmax=\max_k \tilde{h}_k$.
	
\end{lemma}
The row separability of $\bar{\bU}$ is less explicit compared to $\bU$. In addition to depending on the membership matrix $\bZ$, it depends on the row separability of $\bOmega$ and the heterogeneity matrix $\bH$. While $\delta_{\bOmega}>0$ of Lemma~\ref{lem:Urow_2} seems natural, the condition $\delta_{\bH,\bOmega}>0$ of Lemma~\ref{lem:Urow} might be untrue. For instance, we have $\delta_{\bH,\bOmega}=0$ if $h_{j_1}\Omega(k_1,:)=h_{j_2}\Omega(k_2,:)$ for two vertices $j_1\in\calV_{k_1}$ and $j_2\in\calV_{k_2}$ in which case the vertices $j_1$ and $j_2$ are indistinguishable by the eigenvectors of $\bL$. Technically speaking, both $\bU$ and $\bar\bU$, under reasonable conditions, can be used for the clustering of vertices. Our method directly estimates $\bU$ from multiple views of LBM data matrices. 

\begin{rmk}
The matrix $\bOmega$ can be rank deficient under LBM while the DCBM assumes $\bOmega$ to be full rank.  
Under the DCBM, the eigenspace of $\EE\bW$ is equivalent to the eigenspace of $\bH\bZ$, which admits a much simpler separability property such that 
	$
	\|\bar{U}(j_1,:)-\bar{U}(j_2,:)\|_{\ell_2} = \sqrt{h_{j_1}^2/\tilde{h}^2_{k_1}+h^2_{j_2}/\tilde{h}^2_{k_2}},\quad {\rm if}\ j_1\in\calV_{k_1}, j_2\in\calV_{k_2}\ {\rm for}\ k_1\neq k_2$.
\end{rmk}

\def\subbullet{_{\scriptscriptstyle \bullet}}

\subsection{Multi-view Sparse LBM (msLBM)}
We next describe the msLBM framework for learning a consensus knowledge graph using $m$ observed weighted graphs with a common vertex set $\calV = \{v_j\}_{j\in [n]}$ 
\begin{equation}\label{eq:data_D}
\Big\{  \calG_s = \{\calV, \bW_s\}: \calV = \{v_j\}_{j\in [n]}, \bW_s = \big(W_s(j_1,j_2)\big)_{j_1\in[n]}^{j_2\in[n]} \Big \}_{s \in [m]} .
\end{equation}
To learn the consensus graph while accounting for the between-view heterogeneity, we propose the following msLBM 
\begin{equation}
	\label{eq:Ws}
	\bW_s = \bH_s ( \bZ\bOmega\bZ^{\top})  \bH_s^{\top}  + \bTheta_s + \bE_s, \quad s = 1,...,m,
\end{equation}
where both $\bTheta\subbullet\equiv \{\bTheta_s\}_{s \in [m]} $ and $\bOmega$ are assumed to be sparse, 
$\bE_s$ represents the sampling error from the $s\supth$ view. Here $\bC =\bZ \bOmega \bZ^\top$ represents the consensus graph as in Section~\ref{sec:LBM} while the sparse bias term $\bTheta\subbullet$ reflects view-specific patterns, capturing the between view heterogeneity in knowledge graph structure. The view-specific diagonal matrices $\bH\subbullet = \{\bH_s\}_{s \in [m]}$ in (\ref{eq:Ws}) capture the heterogeneity in the information content for the nodes across views. Additionally, the distributions of the error matrices $\{\bE_s\}_{s\in [m]}$ can be different for each view, accommodating the heterogeneity in the noise level.  Here and in the sequel, we use the subscript $``\subbullet"$ to index all views $s \in [m]$.

Rewriting the msLBM model (\ref{eq:Ws}) as
\begin{equation}
	\label{eq:Ws2}
	\bW_s = \bL_s + \bTheta_s +\bE_s \ \mbox{with\ } \bL_s =  \bH_s ( \bZ\bOmega\bZ^{\top})  \bH_s^{\top}, \quad s = 1,\cdots,m,
\end{equation}
we note that each $\bW_s$ can be characterized by a noise-corrupted sum of a low-rank matrix and a sparse matrix. The $m$ views $\bW\subbullet=\{\bW_s\}_{s=1}^m$ share the common knowledge through the correlation matrix $\bC = \bZ\bOmega\bZ^{\top}$ while the individual varying part goes into the sparse component. In a special case when $m=1$ (a single noisy matrix decomposition), the model (\ref{eq:Ws2}) is analogous to the noisy version of robust PCA model \citep[e.g.]{candes2011robust,zhou2011godec}. The msLBM aims to leverage information from multiple resources, which is more challenging.

We impose the following assumption on the noise $\bE_s$, which implies that its entries have zero means, equal variances, and have sub-Gaussian tails. This sub-Gaussian condition is mild, which easily holds under various special and useful distributions. For instance, in the case that $\bE_s$ is also sparse but denser than $\bTheta_s$ and $\bOmega$, as shown in Section \ref{sec:realdata}, 
     we can assume $E_s(j_1,j_2)\stackrel{i.i.d.}{\sim} (1-\pi_s)\cdot {\bf 1}_{ \{x=0 \}}+\pi_s\cdot N(0,\sigma_s^2)$ for a small $\pi_s\in(0,1)$. 

\begin{assump}
 \label{assump:E_s2}  
  For $s=1,\cdots,m$, there exists a $\sigma_s  > 0$ such that $E_s(j_1,j_2)$ are i.i.d. and 
 $$
 \EE E_s(j_1,j_2)=0,\quad {\rm Var}\big\{E_s(j_1,j_2)\big\}=\sigma_s^2, \quad {\rm and}\quad \EE \exp\{t\cdot E_s(j_1,j_2)\}\leq \exp\{t^2\sigma_{s}^2\},\ \forall \; t \in\RR
 $$
 for all $1\leq j_1<j_2\leq n$. 
 \end{assump}

\begin{rmk} Our analysis primarily centers on scenarios where the model (\ref{eq:Ws2}) is correctly specified. Nonetheless, our model can be viewed as a robust alternative to some existing models. Specifically, in instances when $\bTheta_s=\mathbf{0}$, our model includes the stochastic block model \citep{holland1983stochastic}, the he degree corrected stochastic block mode \citep{karrer2011stochastic}, and the random dot product graphs \citep{young2007random} as special cases. The inclusion of additional sparse bias parameters $\bTheta_s$ in our model offers increased flexibility and adaptability. 
\end{rmk}

\textit{Identifiability.} It is well recognized in the low-rank plus sparse matrix/tensor literature \citep{candes2011robust,cai2021generalized} that the low-rank part $\bL_s$ and sparse part $\bTheta_s$ are not identifiable if $\bL_s$ is also very sparse. To ensure the identifiability of msLBM, we assume that the column space of $\bL_s$ is incoherent for all $s=1,\cdots, m$.    A symmetric rank-$r$ matrix $\bL \in \RR^{n \times n}$ with the eigendecomposition of the form $\bU \bD \bU\trans$ with $\bU^{\top}\bU=\bI_r$ is said to be incoherent with constant $\mu_0$ if 
$$
\max_{1\leq j\leq n} \|\be_j^{\top}\bU\|\leq \sqrt{\mu_0r/n},
$$
where $\be_j$ denotes the $j$-th canonical basis vector in $\RR^n$.
\begin{assump}\label{assump:incoh}
There exists a $\kappa_0>1$ so that $\lambda_1(\bC)/\lambda_r(\bC)\leq \kappa_0^{1/2}$ and $\lambda_1(\bH_s^2)/\lambda_n(\bH_s^2)\leq \kappa_0^{1/2}$ for all $s \in [m]$. 
\end{assump}
Basically, Assumption~\ref{assump:incoh} requires the matrices $\bC$ and $\bH_s$ for  $s \in [m]$ to be well-conditioned. 
Interestingly, incoherence can be automatically guaranteed by this assumption. 

\begin{lemma}\label{lem:incoh}
Under Assumption~\ref{assump:incoh}, let $\bU_s$ be the top-$r$ left eigenvectors of $\bL_s$, then $\bU_s$ is incoherent with constant $\kappa_0^2$.
\end{lemma}

\begin{rmk}
 By Lemma~\ref{lem:incoh}, the low-rank matrix $\bL_s$ is incoherent and distinguishable from the sufficiently sparse matrix $\bTheta_s$. As shown in Theorem \ref{thm:W-Theta-err}, in the noiseless case,  if the number of non-zero entries of $\bTheta_s$ is smaller than $n$, the low-rank matrix $\bL_s$ and sparse matrix $\bTheta_s$ are distinguishable. By setting $\sigma_s\equiv 0$, Theorem~\ref{thm:W-Theta-err} implies that $\bL_s$ and $\bTheta_s$ can be exactly recovered. 
It is worth pointing out that this condition is sufficient but not necessary. The model may be identifiable even if the cardinality of $\bTheta_s \gg n$, a scenario we leave for future exploration. In the presence of the noise $\bE_s$, we will show in Theorem~\ref{thm:W-Theta-err} in Section~\ref{sec:theory} that $\bL_s$ can be consistently estimated in the sense that the relative error approaches zero as $n\to\infty$. 
\end{rmk}

\section{Multi-view Consensus Graph Learning}\label{sec:method}

To estimate the model parameters under the msLBM  (\ref{eq:Ws}) with observed $\bW\subbullet$, we first assume the rank $r$ is known, and discuss the estimation of $r$ later. Denote 
$$\calF_{n,r,\kappa_1}:=\Big\{\bA\in\RR^{n\times r}: \|\bA(i,:)\|= 1,\ \forall i\in[n] \textrm{ and } \lambda_1(\bA)\leq \kappa_1^{1/4}\lambda_r(\bA)\Big\}
$$ 
the set of all rank-$r$ well-conditioned correlation matrix, and $\calD_{n,\kappa_1}:=\{\bH=\diag(H(1,1),...,H(n,n)): H(i,i)>0,\ \forall i\in[n] \textrm{ and } \lambda_1(\bH)\leq \kappa_1^{1/2}\lambda_n(\bH)\}$. The constraint on condition number enforces incoherent solutions just as implied by Lemma~\ref{lem:incoh}. We can treat $\kappa_1$ as a tuning parameter satisfying $\kappa_1>\kappa_0$. Our algorithm for estimating $\bZ, \bOmega,\bTheta\subbullet,\bH\subbullet$ includes two key steps. We first obtain
estimates 
\begin{align}\label{eq:obj_ideal}
(\hat\bU, \hat\bH\subbullet, \hat\bTheta\subbullet)&:=\argmin_{\bU\in\calF_{n,r,\kappa_1}, \bH_s\in\calD_{n,\kappa_1}, \bTheta_s\in\RR^{n\times n}} \calL(\bU, \bH\subbullet,\bTheta\subbullet)\notag\\
\textrm {with } \quad \calL(\bU, \bH\subbullet,\bTheta\subbullet)&:=\frac{1}{2}\sum_{s=1}^m \alpha_s\|\bW_s-\bTheta_s-\bH_s\bU\bU^{\top}\bH_s\|_{\rm F}^2+\sum_{s=1}^m \lambda_s\|\bTheta_s\|_{\ell_1}.
\end{align}
Here $\hat\bU \hat\bU\trans$ is an estimate for $\bZ\bOmega\bZ\trans$. 
In the second step, we recover $\bZ$ and $\bOmega$ based on $\bUhat$ via clustering. Here the positive $\alpha_s, \lambda_s$'s are tuning parameters with $\sum_{s=1}^m \alpha_s = 1$ and $\|\cdot\|_{\ell_1}$ norm is used to promote sparse solutions for $\{\bTheta_s\}_{s=1}^m$. The weights $\alpha_s$ can be chosen to reflect the noise levels in $\{\bE_s\}_{s\in [m]}$ and the information content levels $\bH\subbullet$. For example, if the noise levels $\sigma_s$ for $s \in [m]$ are known, a natural choice of $\alpha_s$ is $\alpha_s = \sigma_s^{-2} / \sum_{l=1}^m \sigma_l^{-2}$, which is optimal as shown in Theorem \ref{thm:W-Theta-err}. 

The objective function (\ref{eq:obj_ideal}) is highly non-convex, which is often solvable only locally. In Section \ref{sec:altmin} an alternating minimization algorithm to optimize for (\ref{eq:obj_ideal}) assuming that  good initializations $\hat \bU^{(0)}, \hat\bH^{(0)}\subbullet$ and $\hat \bTheta^{(0)}\subbullet$ have been obtained.  In Section~\ref{sec:init}, we propose a procedure for obtaining a warm start. We detail the clustering algorithm for estimating $\bZ$ and $\bOmega$ in Section \ref{sec:cluster}. A data-driven approach for choosing the tuning parameters is discussed in Section~\ref{tuning}.

\subsection{Alternative Minimization}\label{sec:altmin}

Suppose that we obtain reasonably good initializations $\hat \bU^{(0)}, \hat\bH^{(0)}\subbullet$ and $\hat \bTheta^{(0)}\subbullet$. In Section~\ref{sec:init}, we shall introduce a computationally efficient method for obtaining these initializations. To solve (\ref{eq:obj_ideal}), our algorithm iteratively updates $\hat\bU, \hat\bH_s, \hat\bTheta_s$ by alternating minimization. The detailed implementations of these iterations are presented in Sections~\ref{sec:updateU}, \ref{sec:updateS}, and \ref{sec:updateTheta}. In Section~\ref{sec:inexact_fast}, we introduce a fast but inexact updating algorithm of $\hat\bU$ that scales smoothly to large datasets.

\subsubsection{Estimate low-rank factor $\bU$}\label{sec:updateU}
Suppose that, at $t$-th iteration, provided with $\hat\bTheta^{(t)}\subbullet$ and $\hat\bH^{(t)}\subbullet$, we update $\hat \bU^{(t+1)}$ by solving the following minimization problem:
\begin{align}\label{eq:hatUt+1}
\hat \bU^{(t+1)}=\argmin_{\bU\in\calF_{n,r,\kappa_1}} \sum_{s=1}^m \alpha_s \|\bW_s-\hat\bTheta_s^{(t)}-\hat\bH_s^{(t)}\bU\bU^{\top}\hat\bH_s^{(t)}\|_{\rm F}^2,
\end{align}
which has no closed-form solution. However, problem (\ref{eq:hatUt+1}) can be recast to a weighted low-rank approximation problem. Denoting by $\hat \bh_s^{(t)}\in\RR^{n}$ the diagonal entries of $\hat\bH_s^{(t)}$, we then have 
\begin{align*}
\sum_{s=1}^m &\alpha_s \|\bW_s-\hat\bTheta_s^{(t)}-\hat\bH_s^{(t)}\bU\bU^{\top}\hat\bH_s^{(t)}\|_{\rm F}^2\\
=&\Big< (\bU\bU^{\top})\circ (\bU\bU^{\top}), \sum_{s=1}^m \alpha_s(\hat\bh_s^{(t)}\hat\bh_s^{(t)\top})\circ (\hat\bh_s^{(t)}\hat\bh_s^{(t)\top}) \Big>  -  2\Big<\bU\bU^{\top}, \sum_{s=1}^m\alpha_s (\hat\bh_s^{(t)}\hat\bh_s^{(t)\top})\circ (\bW_s-\hat\bTheta_s^{(t)}) \Big>.
\end{align*}
Then 
\begin{align}\label{eq:hatU_wlra}
\hat\bU^{(t+1)}
=&\argmin_{\bU\in\calF_{n,r,\kappa_1}} \big\| (\bU\bU^{\top})\circ \bX^{(t)}-\bY^{(t)}\big\|_{\rm F}^2,
\end{align}
where $\bX^{(t)}\in \RR_+^{n\times n}$ and $\bY^{(t)}\in\RR^{n\times n}$ satisfy
$$\bX^{(t)}\circ \bX^{(t)}=\sum_{s=1}^m \alpha_s(\hat\bh_s^{(t)}\hat\bh_s^{(t)\top})\circ (\hat\bh_s^{(t)}\hat\bh_s^{(t)\top}), \quad\mbox{and}\quad \bX^{(t)}\circ \bY^{(t)}= \sum_{s=1}^m\alpha_s (\hat\bh_s^{(t)}\hat\bh_s^{(t)\top})\circ (\bW_s-\hat\bTheta_s^{(t)}).$$ 
The optimization in (\ref{eq:hatU_wlra}) can be solved as a weighted low-rank approximation problem (WLRA) via existing algorithms including the gradient descent algorithm and EM procedure \citep{srebro2003weighted}. The upper bound of the condition number, $\kappa_1$, in (\ref{eq:hatU_wlra}) is a tuning parameter which addresses regularity concerns.

\subsubsection{Estimate $\bH_s$}\label{sec:updateS}
Provided with $\hat \bU^{(t+1)}\hat\bU^{(t+1)\top}$ and $\hat\bTheta_s^{(t)}$ at the $t$-th iteration, we can estimate $\bH_s$ by minimizing (\ref{eq:obj_ideal}), which is equivalent to 
\begin{equation}
\label{eq:Ns}
    \hat \bH_s^{(t+1)} = \argmin_{\bH\in \calD_{n,\kappa_1}} \|\bW_s - \hat\bTheta_s^{(t)} - \bH (\hat\bU^{(t+1)}\hat\bU^{(t+1)\top}) \bH\|_{\rm F}^2 \quad \text{ for } \; s \in [m] \,.
\end{equation}
The problem (\ref{eq:Ns}) is a weighted rank-$1$ approximation of $\bW_s-\hat\bTheta_s^{(t)}$, which 
generally has no closed-form solution. We propose to use an alternative direction method of multipliers (ADMM) type algorithm to solve the problem (\ref{eq:Ns}). By decoupling the two $\bH$'s in (\ref{eq:Ns}), we write 
\begin{align} \label{eq:Ns2}
    \min_{\bH_1,\bH_2 \in \calD_{n,\kappa_1}} & \; \|\bW_s - \hat \bTheta_s^{(t)} - \bH_1 (\hat\bU^{(t+1)}\hat\bU^{(t+1)\top}) \bH_2\|_{\rm F}^2 \quad
    {\rm s.t.}  \; \bH_1 = \bH_2 \, .
 \end{align}
Problem (\ref{eq:Ns2}) becomes easy when fixing either one of $\bH_1$ and $\bH_2$. Toward that end, 
we propose Algorithm~\ref{ReH} to solve the problem (\ref{eq:Ns2}). Note that we set in Algorithm~\ref{ReH} the input $\widetilde\bW = \bW_s - \hat \bTheta_s^{(t)}$, $\widetilde\bC = \hat\bU^{(t+1)}\hat \bU^{(t+1)\top}$. The output of Algorithm~\ref{ReH} is the estimate $\hat \bH_s^{(t+1)}$.

\begin{algorithm}
\caption{${\rm ReH}(\widetilde\bW,\widetilde\bC,\bH_0)$}
\label{ReH}
\begin{algorithmic}[1]
    \State Input: $\widetilde\bW=\bW_s-\hat\bTheta_s^{(t)}$, the estimated correlation matrix $\widetilde\bC=\hat\bU^{(t+1)}\hat\bU^{(t+1)\top}$, an initialization $\bH_0=\hat\bH_s^{(t)}$;  the maximal iterations ${\rm iter}_{\max}$ and the tolerance parameter $\epsilon_{tol}>0$;condition number $\kappa_1$. 
    
    \State Set $\bH_1 = \bH_2 = \bH_0$, $\lambda=1$ and $t=0$
    
	\While{$t < {\rm iter}_{\max}$}
			\State $t=t+1$
			\State ${\rm diag}(\bH_1)_j = \frac{\sum_{h=1}^n \widetilde{W}_{jh}\widetilde{C}_{jh} {\rm diag}(\bH_2)_h + \lambda {\rm diag}(\bH_2)_j}{\sum_{h=1}^n (\widetilde{C}_{jh} {\rm diag}(\bH_2)_h)^2 + \lambda}$ 
			
			\State ${\rm diag}(\bH_2)_j = \frac{\sum_{h=1}^n \widetilde{W}_{jh}\widetilde{C}_{jh} {\rm diag}(\bH_1)_h + \lambda {\rm diag}(\bH_1)_j}{\sum_{h=1}^n (\widetilde{C}_{jh} {\rm diag}(\bH_1)_h)^2 + \lambda}$ 
			
	        \State \textbf{If} $\|\bH_1 - \bH_2\|_{\rm F} \leq \epsilon_{tol}$ \textbf{then} break
            
            \State $\lambda = \lambda + 1$  
			\EndWhile
   \State Choose the smallest $\delta>0$ such that the condition number of 
   $\hat\bH_s^{(t+1)} = \frac{\bH_1 + \bH_2}{2}+\delta \bI$ is smaller than $\kappa_1$.
    \State Output: $\hat \bH_s^{(t+1)}$.
\end{algorithmic}
\end{algorithm}

We remark that the parameter $\lambda$ in Algorithm~\ref{ReH} is for regularization, commonly used in ADMM type algorithms. It is empirically important because some entries of $\bH_1$ and $\bH_2$ are often small, making the algorithm unstable on large-scale computations. The regularization $\delta \bI$ added in the last step enforces a restriction on the condition number, ensuring it remains bounded. The parameter $\kappa_1$ needs to be larger than $\kappa_0$. In practice, an estimate of $\kappa_0$ is obtained using the individual data matrix $\bW_s$. For all $s\in[m]$, we find the best rank-$r$ approximation of $\bW_s$, diagonally scale it to a correlation matrix, and estimate $\kappa_0$ using the largest one of the condition numbers of these correlation matrices.

\subsubsection{Estimate sparse individual component $\bTheta_s$}\label{sec:updateTheta}
Finally, provided with $\hat \bU^{(t+1)} \hat \bU^{(t+1)\top}$ and $\hat \bH_s^{(t+1)}$ at the $t$-th iteration, we can estimate the sparse individual component $\bTheta_s$ by solving
\begin{align}\label{eq:Thetas}
\hat\bTheta_s^{(t+1)}:=\argmin_{\bTheta\in \RR^{n\times n}} \frac{\alpha_s}{2} \|\bW_s-\bTheta-\hat\bH_s^{(t+1)}\hat \bU^{(t+1)}\hat \bU^{(t+1)\top}\hat\bH_s^{(t+1)}\|_{\rm F}^2+\lambda_s \|\bTheta\|_{\ell_1} \quad \text{ for } \; s \in [m] \,.
\end{align}
Problem (\ref{eq:Thetas}) has a closed-form solution through a simple entry-wise soft-thresholding method. To this end, we propose Algorithm~\ref{ReTheta} to obtain $\hat\bTheta_s^{(t+1)}$ where the threshold is set at $\tau_s=\lambda_s/\alpha_s$. 

\begin{algorithm}[ht]
\caption{${\rm Re}\Theta(\bW_s,\hat \bH_s^{(t+1)}, \hat\bC^{(t+1)}=\hat\bU^{(t+1)}\hat \bU^{(t+1)\top}, \tau_s)$}
\label{ReTheta}
\begin{algorithmic}[1]
    \State Input: $\bW_s, \hat\bH_s^{(t+1)}$; the estimated correlation matrix $\hat\bC^{(t+1)}=\hat\bU^{(t+1)}\hat\bU^{(t+1)\top}$; the threshold $\tau_s=\lambda_s/\alpha_s$.
    
    \State $\widetilde\bDelta_s = \bW_s - \hat\bH_s^{(t+1)} \hat\bC^{(t+1)} \hat\bH_s^{(t+1)}$.
    
    \State Get $\hat\bTheta_s^{(t+1)}$ by $(\hat\bTheta_s^{(t+1)})_{ij} = \begin {cases} 
(\widetilde\bDelta_s)_{ij}-\tau_s,  &\textrm{ if }(\widetilde\bDelta_s)_{ij}> \tau_s\\
0, &\textrm{ if } (\widetilde\bDelta_s)_{ij}\in[-\tau_s, \tau_s]\\
(\widetilde\bDelta_s)_{ij}+\tau_s,&\textrm{ if } (\widetilde\bDelta_s)_{ij}<- \tau_s
\end{cases}$
    
    \State Output: $\hat\bTheta_s^{(t+1)}$
\end{algorithmic}
\end{algorithm}
Putting together the iterative rules in Section~\ref{sec:updateU}, \ref{sec:updateS} and \ref{sec:updateTheta}, we solve the problem~(\ref{eq:obj_ideal}) by Algorithm~\ref{algo:exact}.

\begin{algorithm}
\caption{Alternating Minimization for Solving (\ref{eq:obj_ideal})}\label{algo:exact}
\begin{algorithmic}[2]
\State Input: $\bW_s$, the weight and regularization parameters $\alpha_s, \lambda_s, s\in[m]$; the rank $r$; the maximal iterations ${\rm iter}_{\max}$ and the tolerance parameter $\epsilon_{tol}>0$.
			
\State Warm initialization: $\hat \bC^{(0)}=\hat\bU^{(0)}\hat\bU^{(0)\top}$, $\hat\bTheta_s^{(0)}, \hat\bH_s^{(0)}, s\in[m]$. 
			
\State Set the step counter $t = 0$
\While{$t < {\rm iter}_{\max}$}
			
			\State $t=t+1$
			
			\State Use weighted low-rank approximation of (\ref{eq:hatU_wlra}) to update
			$\hat \bU^{(t)}$ and set $\hat\bC^{(t)}=\hat\bU^{(t)}\hat\bU^{(t)\top}$;
			
			\State Use  Algorithm \ref{ReH} to update 
			$\hat \bH_s^{(t)} = {\rm ReH}\big(\bW_s - \hat\bTheta_s^{(t-1)}, \hat\bC^{(t)}, \hat \bH_s^{(t-1)}\big)$ for $s\in[m]$;
			
			\State Use Algorithm \ref{ReTheta} to update 
		        $
		        \hat\bTheta_s^{(t)} = {\rm Re}\Theta(\bW_s, \hat \bC^{(t)}, \hat \bH_s^{(t)}, \lambda_s/\alpha_s)$ for $s\in[m]$;

	        \State \textbf{If} $\|\hat\bC^{(t)} - \hat\bC^{(t-1)}\|_{\rm F} \leq \epsilon_{tol}$ \textbf{then} break
        
			\EndWhile
			
			\State Output: $\hat \bC^{(t)}$ and $\hat\bTheta_s^{(t)}, \hat\bH_s^{(t)}, s\in[m]$.

\end{algorithmic}
\end{algorithm}

\subsubsection{Inexact but faster update of $\bU$}\label{sec:inexact_fast}
While the proposed update of $\hat\bU^{(t+1)}$ via problem (\ref{eq:hatUt+1}) is (at least locally) polynomial-time solvable by gradient descent, it is still quite slow on large-scale real datasets, e.g.,  the clinical knowledge graph example in Section~\ref{sec:realdata}. We observe that a simple but fast inexact update of $\hat\bU^{(t+1)}$ yields favorable performances. 

The major computation bottleneck of the problem (\ref{eq:hatUt+1}) is the sum of matrix Frobenius norms which does not admit a closed-form solution. However, the optimization problem for each matrix Frobenius norm in (\ref{eq:hatUt+1}) becomes easy. 
For a fixed $s$ and given $\bW_s, \hat\bTheta_s$, the solution of 
\begin{align}\label{eq:hatU_s}
(\widetilde\bH_s^{(t)}, \hat\bU_s^{(t)}):=\argmin_{\bU\in\calF_{n,r,\kappa_1},\bH\in\calD_{n,\kappa_1}} \|\bW_s-\hat\bTheta_s^{(t)}-\bH\bU\bU^{\top}\bH\|_{\rm F}^2
\end{align}
is attainable by a truncated eigenvalue decomposition. Indeed, observe that the solution to the problem (\ref{eq:hatU_s}) amounts to a best rank-$r$ approximation of $\bW_s-\hat\bTheta_s$ by a positive semi-definite matrix, which is attainable by a truncated eigenvalue decomposition as described in Algorithm~\ref{algo:hatUs}.  As a result, we can simply obtain $(\hat\bU_s^{(t)}\hat\bU_s^{(t)\top}, \widetilde\bH_s^{(t)})={\rm ReC}(\bW_s-\hat\bTheta_s^{(t)}, r)$.

\begin{algorithm}
\caption{${\rm ReC}(\bW,r)$}\label{algo:hatUs}
\label{ReC}
\textbf{Input}: the symmetric matrix $\bW$ and rank $r$.
\begin{algorithmic}[2]    
   \State Compute the eigen-decomposition: $\bW=\sum_{i=1}^n \lambda_i \bv_i\bv_i^{\top}$ and denote $\tilde\lambda_i=\lambda_i\cdot\mathbbm{1}(\lambda_i>0)$. 

    \State Get the best rank-$r$ approximation of $\widetilde\bW=\sum_{i=1}^r\tilde \lambda_i \bv_i\bv_i^{\top}$ $=:\bV \bSigma \bV^\top$. 
    
    \State Get $\bU = \bV \bSigma^{\frac{1}{2}}$ and its row normalization matrix $\tilde{\bU}$. The $i$ row of  $\tilde{\bU}$ is  $\tilde{\bU}_{i:} = \frac{1}{\|\bU_{i:}\|_2}\bU_{i:}$. 
    
    \State $\bC = \tilde{\bU} \tilde{\bU}^{\top}$, $\bH = {\rm diag}(\|\bU_{1:}\|_2, \dots,\|\bU_{n:}\|_2)$.
    
    \State Output: $\tilde{\bU}\tilde{\bU}^{\top}$.
\end{algorithmic}
\end{algorithm}
Unlike (\ref{eq:hatU_s}), problem (\ref{eq:hatUt+1}) involves the sum of multiple low-rank approximations, which admits no closed-form solution. To speed up the update of $\hat\bU^{(t+1)}$, we turn to solve the individual problem (\ref{eq:hatU_s}) for each $s$ independently, and then to update $\hat\bU^{(t+1)}$ as a weighted average of the correlation matrix $\hat\bU_s^{(t)}\hat\bU_s^{(t)\top}$ estimated locally on each $\bW_s$. Then, we calculate the weighted average
\begin{align}\label{eq:tildeC}
\widetilde \bC^{(t)}=\sum_{s=1}^m \frac{\alpha_s}{\sum_s \alpha_s}\hat \bU_s^{(t)}\hat \bU_s^{(t)\top}. 
\end{align}
Since all $\alpha_s>0$, it is easy to check that $\widetilde \bC^{(t)}$ is indeed a correlation matrix, i.e., $\widetilde \bC^{(t)}$ is positively semi-definite and all diagonal entries equal $1$. However, the rank of $\widetilde \bC^{(t)}$ is larger than $r$.  
Finally, by applying Algorithm~\ref{algo:hatUs} on $\widetilde \bC^{(t)}$, we obtain the final update by $\hat\bC^{(t+1)}=\hat\bU^{(t+1)}\hat\bU^{(t+1)\top}={\rm ReC}(\widetilde\bC^{(t)},r)$. 

Equipped with this fast and inexact update of $\hat\bU^{(t+1)}$, our estimating procedure on the large-scale dataset is summarized in Algorithm~\ref{algo:inexact_all}.

\begin{algorithm}[ht]
	\caption{Alternating minimization with inexact update}
	\label{algo:inexact_all}
	\begin{algorithmic}[3]
			\State Input: $\bW_s$, the weight and regularization parameters $\alpha_s, \lambda_s, s\in[m]$; the rank $r$; the maximal iterations ${\rm iter}_{\max}$ and the tolerance parameter $\epsilon_{tol}>0$.
			
			\State Warm initialization: $\hat \bC_s^{(0)}=\hat\bU_s^{(0)}\hat\bU_s^{(0)\top}$, $\hat\bTheta_s^{(0)}, \hat\bH_s^{(0)}, s\in[m]$. 
			
			\State Set the step counter $t = 0$ and $q_s=\alpha_s(\sum_{s=1}^m \alpha_s)^{-1}$; 
			\While{$t < {\rm iter}_{\max}$}
			
			\State $t=t+1$
			
			\State Use Algorithm \ref{ReC} to update
			$$
			\hat \bC^{(t)} := \hat\bU^{t}\hat\bU^{(t)\top}= {\rm ReC}\Big(\sum_{s=1}^m q_s \hat\bC_s^{(t-1)},r\Big),\quad s\in[m]
			$$ 
			
			\State \textbf{If} $\|\hat\bC^{(t)} - \hat\bC^{(t-1)}\|_{\rm F} \leq \epsilon_{tol}$ \textbf{then} break
			
			\State Use  Algorithm \ref{ReH} to update 
			$$
			\hat \bH_s^{(t)} = {\rm ReH}\big(\bW_s - \hat\bTheta_s^{(t-1)}, \hat\bC^{(t)}, \hat \bH_s^{(t-1)}\big),\quad s\in[m]
			$$
			
			\State Use Algorithm \ref{ReTheta} to update 
		        $$
		        \hat\bTheta_s^{(t)} = {\rm Re}\Theta(\bW_s, \hat \bC^{(t)}, \hat \bH_s^{(t)}),\quad s\in[m]
		        $$  
			
			\State Use Algorithm \ref{ReC} to update
			$$
			\hat\bC_s^{(t)} =  {\rm ReC}(\bW_s - \hat\bTheta_s^{(t)}, r),\quad s\in[m]
			$$
        
			\EndWhile
			
			\State Output: $\hat \bC^{(t)}$ and $\hat\bTheta_s^{(t)}, \hat\bH_s^{(t)}, s\in[m]$.
	\end{algorithmic} 
\end{algorithm}

\def\supzero{^{(0)}}

\subsection{Warm Initialization}\label{sec:init}

We next describe a procedure for obtaining $\hat\bU\supzero$, $\hat\bH\supzero\subbullet$, and $\hat\bTheta\supzero\subbullet$ as warm initializations of 
the iterative algorithm discussed in Section \ref{sec:altmin}. 
Recall the equation (\ref{eq:Ws2}) that amounts to a low-rank plus sparse decomposition of each $\bW_s$. We follow the penalized method in \cite{tao2011recovering} to estimate the low-rank matrix $\bL_s=\bH_s\bU\bU\trans\bH_s$ and the sparse matrix $\bTheta_s$:
\begin{equation}  \label{eq:asalm}
    \min_{\bL_s,\bTheta_s\in \RR^{n\times n}} \;  \frac{1}{2}\|\bW_s-\bL_s-\bTheta_s\|_{\rm F}^2+\mu\|\bL_s\|_\ast + \tau \|\bTheta_s\|_1,
\end{equation}
where the nuclear norm $\|\cdot\|_{\ast}$ promotes low-rank solution and $\|\cdot\|_{\ell_1}$ norm promotes sparse solution. The parameters $\mu,\tau>0$ control the rank and sparsity. The problem (\ref{eq:asalm}) is convex and \cite{tao2011recovering} proposed an alternating splitting augmented Lagrangian method (ASALM) to solve it. 
Their key idea is to reformulate (\ref{eq:asalm}) into the following favorable form:
\begin{equation} \label{eq:asalm2}
\begin{split}
    \min_{\bL_s,\bTheta_s,\bE_s\in\RR^{n\times n}} & \;  \frac{1}{2}\|\bE_s\|_{\rm F}^2+\mu\|\bL_s\|_\ast + \tau \|\bTheta_s\|_{\ell_1} \\
    {\rm s.t.}\ & \; \bL_s + \bTheta_s + \bE_s = \bW_s.
\end{split}
 \end{equation}
The augmented Lagrangian function of (\ref{eq:asalm2}) is
\begin{equation} \label{0.3}
\begin{split}
    \mathcal{L}(\bL_s,\bTheta_s,\bE_s,\bLambda,\beta)& := \frac{1}{2}\|\bE_s\|_{\rm F}^2+\mu\|\bL_s\|_\ast + \tau \|\bTheta_s\|_{\ell_1} \\
    & - \left \langle \bLambda, \bL_s + \bTheta_s + \bE_s - \bW_s \right \rangle
    + \frac{\beta}{2} \|\bL_s + \bTheta_s + \bE_s - \bW_s\|_{\rm F}^2,
\end{split}
\end{equation}
where $\beta>0$ is a tuning parameter. The iterative scheme of ASALM then consists of the following updates with explicit solutions at the $k$-th iteration:
\begin{equation}\label{eq:ASALM_iter}
\left\{
             \begin{array}{l}
             \bE_s^{(k+1)} \in \argmin_{\bE_s \in \RR^{n \times n}} \frac{1}{2} \|\bE_s\|_{\rm F}^2 + \frac{\beta}{2}\|\bE_s+\bL_s^{(k)} + \bTheta^{(k)}_s - \frac{1}{\beta} \bLambda^{(k)} - \bW_s\|_{\rm F}^2  \\
             \bTheta_s^{(k+1)} \in \argmin_{\bTheta_s \in \RR^{n \times n}} \tau \|\bTheta_s\|_{\ell_1} + \frac{\beta}{2}\|\bTheta_s + \bL_s^{(k)} +  \bE_s^{(k+1)} - \frac{1}{\beta} \bLambda^{(k)} -\bW_s\|_{\rm F}^2 \\
             \bL_s^{(k+1)} \in \argmin_{\bL_s \in \RR^{n \times n}} =\mu\|\bL_s\|_\ast + \frac{\beta}{2}\| \bL_s + \bTheta_s^{(k+1)} +  \bE_s^{(k+1)} - \frac{1}{\beta} \bLambda^{(k)} -\bW_s\|_{\rm F}^2 \\
             \bLambda^{(k+1)} = \bLambda^{(k)} - \beta(\bL_s^{(k+1)} + \bTheta_s^{(k+1)} + \bE_s^{(k+1)} - \bW_s) 
             \end{array} .
\right.
\end{equation}
It is straightforward to see that $\bE_s^{(k+1)}$ has a closed-form solution and the solution $\bTheta_s^{(k+1)}$ is attainable by entry-wise thresholding. Explicit solutions for $\bTheta_s^{(k+1)}$ and $\bL_s^{(k+1)}$ can be obtained as 
\begin{align*}
\bTheta_s^{(k+1)} & =\calS_{\tau\beta^{-1}}(\bW_s+\beta^{-1}\bLambda^{(k)}-\bE_s^{(k+1)}-\bL_s^{(k)}) \\
\mbox{and}\quad \bL_s^{(k+1)} & =\calQ_{\mu\beta^{-1}}(\bW_s+\beta^{-1}\bLambda^{(k)}-\bE_s^{(k+1)}-\bTheta_s^{(k+1)}),
\end{align*}
respectively, where for any $a>0$ and matrix $\bM$ with singular value decomposition (SVD)
$\bU \bSigma \bV^\top$, 
\begin{align}
   & (\mathcal{S}_a(\bM))_{ij} := \max \{|M_{ij}|- a,0 \}\cdot {\rm sign}(M_{ij}). \label{eq:Salpha_def} \\
   & \mathcal{Q}_a(\bM):= \bU \mathcal{S}_a(\bSigma) \bV^\top .\label{eq:Dalpha_def}
\end{align}
Therefore, ASALM for (\ref{eq:asalm2}) updates $(\bL_s^{(k+1)}, \bTheta_s^{(k+1)}, \bE_s^{(k+1)})$ via the following computations in Algorithm~\ref{algo:ASALM}.
\begin{algorithm}[ht]
	\caption{The k-th iteration of the extended ASALM for (\ref{eq:asalm2}):} \label{algo:ASALM}
	\begin{algorithmic}[2]
			\State Compute $\bE_s^{(k+1)} = \frac{\beta}{1+\beta}(\bW_s+\beta^{-1}\bLambda^{(k)} - \bL_s^{(k)} - \bTheta_s^{(k)})$.
			\State Compute $\bTheta_s^{(k+1)} = \calS_{\tau\beta^{-1}}(\bW_s+\beta^{-1}\bLambda^{(k)}-\bE_s^{(k+1)}-\bL_s^{(k)})$.
			\State Compute $\bL_s^{(k+1)}= \calQ_{\mu\beta^{-1}}(\bW_s+\beta^{-1}\bLambda^{(k)}-\bE_s^{(k+1)}-\bTheta_s^{(k+1)})$.
			\State Update $\bLambda^{(k+1)} = \bLambda^{(k)} - \beta( \bL_s^{(k+1)}  + \bTheta_s^{(k+1)} + \bE_s^{(k+1)} - \bW_s)$.
	\end{algorithmic} 
\end{algorithm}

The initialization $\hat\bL_s^{(0)}$ is then passed to Algorithm~\ref{algo:hatUs} which outputs $\hat\bH_s^{(0)}$ and $\hat\bU_s^{(0)}$. Then we ensemble $\hat\bU_s^{(0)}$ via (\ref{eq:tildeC}) and Algorithm~\ref{algo:hatUs} to generate the initial estimate $\hat\bU^{(0)}$. Note that the ensemble  (\ref{eq:tildeC}) relies on the weight $\alpha_s$ assigned to the $s\supth$ view. We shall discuss the way to choose these weights in Section~\ref{tuning}.

\begin{rmk}
Algorithm \ref{algo:exact} tackles a highly non-convex estimation problem, making theoretical proof of its convergence challenging. While the initial estimators demonstrate convergence \citep{tao2011recovering}, they do not achieve the optimal rate of $O(1/m)$ associated with multi-view learning since they are trained individually for each view. Nevertheless, we posit that with a warm initialization, Algorithm \ref{algo:exact} is likely to yield estimators that are superior to the initial ones. This assertion is supported by our simulation studies in Section \ref{sec:simulations}  and real data analysis in Section \ref{sec:realdata}. When additional validation data such as those from knowledge graph in the real data example, one may mitigate the convergence to a local optimum by monitoring the algorithm  performance at each iteration, as elaborated in Section \ref{tuning}.
\end{rmk}

\subsection{Clustering and Network Analysis}\label{sec:cluster}

In the final step, we apply the K-means algorithm \citep{steinhaus1956division} on $\hat\bU \in \calF_{n,r}$  from the output of Algorithm~\ref{algo:exact}  to recover $\bZ$ and $\bOmega$. The K-means algorithm aims to solve the optimization problem
\begin{align}\label{eq:Kmeans}
(\widehat \bZ, \widehat \bK)=\argmin_{\bZ\in\scrZ_{n,K}, \bK\in\RR^{K\times r}} \|\bZ\bK-\widehat\bU\|_{\rm F}^2
\end{align}
where the $k$-th row of $\bK$ represents the $k$-th centroid in the $r$-dimensional space. Even though the exact solution to the optimization problem in (\ref{eq:Kmeans}) is generally NP-hard \citep{mahajan2009planar}, there exist efficient algorithms to find an approximate solution whose objective value is within a constant fraction of the global minimal value  \citep{kumar2004simple,awasthi2015hardness}. Therefore, given $\eps\in(0,1)$, we calculate the $(1+\eps)$-approximate solution:
\begin{align}\label{eq:app_Kmeans}
&\quad (\widehat \bZ, \widehat \bK)\in \scrZ_{n,K}\times \RR^{K\times r}\nonumber\\
{\rm s.t.}\quad& \|\widehat\bZ\widehat\bK-\widehat\bU\|_{\rm F}^2\leq (1+\eps)\min_{\bZ\in\scrZ_{n,K}, \bK\in\RR^{K\times r}} \|\bZ\bK-\widehat\bU\|_{\rm F}^2.
\end{align}
Although the solutions to (\ref{eq:app_Kmeans}) might not be unique, they all attain the same theoretical guarantees. 
We denote by $\widehat \bZ$ any output from the optimization of (\ref{eq:app_Kmeans}). The {\it group-level weight matrix } $\bOmega$ can be naturally estimated by
\begin{equation}\label{eq: Omegahat}
\widehat \bOmega= (\widehat\bZ^{\top}\widehat\bZ)^{-1}\widehat \bZ^{\top} \widehat \bC  \widehat\bZ (\widehat\bZ^{\top}\widehat\bZ)^{-1}.
\end{equation} 
\begin{rmk}
Since the matrix $\widehat\bU$ has rank $r$, $\widehat\bOmega$ has rank at most $r$. If the underlying graph is sparse, a hard thresholding procedure can be applied on $\widehat\bOmega$ to obtain its sparsified version. In Section~\ref{sec:realdata}, we will show the sparsity of $\widehat\bOmega$ in the real data analysis. 
\end{rmk}

\subsection{Tuning Parameters}
\label{tuning}

In our algorithm, several tuning parameters require careful selection, including $\alpha_s$, $\lambda_s$, the rank $r$ for the objective equation \eqref{eq:obj_ideal}, and the number of groups $K$, 
as well as $\mu$, $\tau$, and $\beta$ for the initialization process (\ref{eq:asalm}). To optimize computational efficiency, we initially determine the parameters $\mu$, $\tau$, $\beta$, and $r$ through a grid search. Subsequently, we select $\alpha_s$ and $\lambda_s$.

Specifically, for the convex optimization problem (\ref{eq:asalm}), we adopt the guidelines from \cite{tao2011recovering} to set the parameters: $\mu$ is set as $\sqrt{n^2 + \sqrt{8} n} \delta$, $\tau$ as $n^{-1/2}$, and $\beta$ as $a n^2/\|\bW_s\|_{\ell_1}$ for each $s \in [m]$. The values of $\delta$, $a$, and the rank $r$ are fine-tuned using a grid search.  For instance, if we have access to labels on a subset of concept pairs with their relatedness scores manually annotated (as in Section \ref{sec:realdata}), we can calculate the Spearman's rank correlation between the labeled relatedness scores and the corresponding estimated entries from $\sum_{s=1}^m \hat\bU_s^{(0)} (\hat\bU_s^{(0)})^{\top}$.
The tuning parameters that result in the highest rank correlation are then chosen. For computational efficiency, a preliminary estimate of the rank $r$ can also be constructed based on the eigenvalue decay of $\bW_s$, which is a widely recognized technique for determining the rank of low-rank matrices \citep{jolliffe2005principal}.

We estimate $\sigma_s^2$ as $\hat\sigma_s^2=\|\bW_s-\hat\bL_s^{(0)} - \hat\bTheta_s^{(0)}\|_{\rm F}^2/n^2$. The weight parameters are then set as $\alpha_s = c_s \hat h_s^{-4}\hat\sigma_s^{-2}$ and the regularization parameters as $\lambda_s = c \alpha_s \hat \sigma_s \log^{1/2}n$, where $\hat h_s$ represents the average of the diagonal entries of $\hat \bL_s^{(0)}$. The constants $c_s$ and $c$ are determined again through grid search. The theoretical underpinnings for the rates of $\alpha_s$ and $\lambda_s$ are explained in Section \ref{sec:theory}.

To select the number of groups $K$ without prior grouping information, we recommend using the elbow method, which involves plotting the within-cluster sum of squared errors (WSS) against various $K$ values, and the Silhouette method. However, if some group labels are available, the sum of normalized mutual information (NMI) and adjusted Rand index (ARI) can be utilized. When only partial labels are available, such as pairs within and between groups, we suggest using a composite score defined as the sum of sensitivity and specificity to ascertain the optimal $K$. The procedure of selecting these tuning parameters is further illustrated in Sections \ref{sec:simulations} and \ref{sec:realdata}.

\section{Theory}\label{sec:theory}

In this section, we provide theoretical analyses of the performance of the estimator (\ref{eq:obj_ideal}) under the msLBM model with Assumptions~\ref{assump:lowrank}-\ref{assump:incoh}. 

\subsection{Joint Estimation Bounds for Weight and Heterogeneity Matrix}

Let $\kappa_1>\kappa_0$ be the condition number used in (\ref{eq:obj_ideal}) and $\hat \bU$, $\{\hat \bH_s\}_{s=1}^m, \{\hat\bTheta_s\}_{s=1}^m$ be the estimators. For $\forall s\in[m]$, denote $\hat\bDelta_s={\red \hat\bL_s-\bL_s}:=\hat\bH_s\hat \bU\hat \bU^{\top}\hat \bH_s-\bH_s\bU\bU^{\top}\bH_s$. Denote $\Psi_s={\rm supp}(\bTheta_s)$ the support of $\bTheta_s$, i.e., the locations of non-zero entries of $\bTheta_s$. The joint estimation bounds for $\{\hat\bDelta_s\}_{s=1}^m$ in both Frobenius and Sup norms are as follows.  

\begin{theorem}\label{thm:W-Theta-err}
Under Assumptions~\ref{assump:lowrank}-\ref{assump:incoh}, there exist constants $C_0', C_0,C_1,\cdots,C_8>0$ depending only on $\kappa_1$ such that if $m\leq C_0'r$, $C_0r^2|\Psi_s|\leq n$ and $\lambda_s\asymp \alpha_s \sigma_s\log^{1/2}n$ is appropriately chosen, we get with probability at least $1-mn^{-2}$ that
\begin{align*}
\sum_{s=1}^m \alpha_s \|\hat\bDelta_s\|_{\rm F}^2\leq&C_1 rn\cdot \big(\max_{s\in[m]} \alpha_s \sigma_s^2\big)+C_2\sum_{s=1}^m \alpha_s \sigma_s^2|\Psi_s|\log n\\
\sum_{s=1}^m\alpha_s\|\hat\bDelta_s\|_{\ell_\infty}^2\leq&  C_3r^3\cdot  \big(\max_{s\in[m]} \alpha_s \sigma_s^2\big)+C_4\sum_{s=1}^m \alpha_s \sigma_s^2r^2\cdot \frac{|\Psi_s|\log n}{n}
\end{align*}
and
\begin{align*}
\sum_{s=1}^m\alpha_s \|\hat \bTheta_s-\bTheta_s\|_{\rm F}^2\leq&\max_{s\in[m]}|\Psi_s|\cdot C_5r^3 \big(\max_{s\in[m]} \alpha_s \sigma_s^2\big)+\max_{s\in[m]}|\Psi_s|\cdot C_6\sum_{s=1}^m \alpha_s\sigma_s^2\log n\\
\sum_{s=1}^m\alpha_s \|\hat \bTheta_s-\bTheta_s\|_{\ell_\infty}^2\leq& C_7r^3\cdot  \big(\max_{s\in[m]} \alpha_s \sigma_s^2\big)+C_{8}\sum_{s=1}^m\alpha_s\sigma_s^2\log n.
\end{align*}
\end{theorem}

The bounds established in Theorem~\ref{thm:W-Theta-err} require no conditions on the magnitudes of the non-zero entries of the heterogeneity matrices $\bTheta_s$'s. This is due to the penalty by the $\ell_1$-norm. On the other hand, to prove sharp bounds, we require that the support sizes of $\bTheta_s$ are upper bounded by $O(n)$. This is a mild condition since the matrix is indeed very sparse on EHR datasets. See the real data example in Section~\ref{sec:realdata}. The probability bound $1-mn^{-2}$ can be improved to $1-(mn)^{-2}$ if we replace the $\log n$ term in the upper bounds by $\log (mn)$. 

\begin{rmk}
The view-wise heterogeneity matrix $\bH_s$ significantly enhances model flexibility. This heterogeneity is a crucial methodological innovation, enabling superior performance with real data. However, this comes at a cost of an increased number of model parameters, with $O(mn)$ attributable to estimating $\{\bH_s\}$.  In our theoretical analysis, we concentrate on the case where $m=O(r)$, which strikes a balance between maximizing model flexibility and minimizing model complexity.  This assumption is well-suited to real data applications. For example, in our real data experiment, the rank $r$ is approximately 250, while $m \leq 10$.
The condition $m=O(r)$ also allows for technical convenience.  
The main parameter of interest, the shared correlation matrix $\bC$, has a model complexity of $O(rn)$. If $m \gg r$, the model complexity is dominated by $O(mn)$, and accurately estimating the heterogeneity matrices becomes a major bottleneck in fitting the msLBM. 
When $m\gg r$, additional structure on heterogeneity is needed in order to better borrow information across views. For example, one may assume some of the views share similar $\bH_s$. This can significantly reduce the degree of heterogeneity in $\{\bH_s\}_{s=1}^m$. The estimation procedures can be modified accordingly, and we expect that parallel theoretical results can be derived.
\end{rmk}

By choosing the weight $\alpha_s\asymp m^{-1}$, Theorem~\ref{thm:W-Theta-err} implies that w.h.p., 
\begin{align}\label{eq:hatDeltas_ineq2}
\sum_{s=1}^{m}\|\hat\bDelta_s\|_{\rm F}^2=O\Big(rn\cdot \max_{s\in[m]} \sigma_s^2+\sum_{s=1}^m \sigma_s^2|\Psi_s|\log n\Big).
\end{align}
 The degrees of freedom of the parameters $\{\bH_s\}_{s=1}^m$  and $\bU\bU^{\top}$ in msLBM model is $O\big(rn\big)$. It implies that the first term in the RHS of (\ref{eq:hatDeltas_ineq2}) is sharp if $\sigma_s^2\asymp \sigma^2$ for all $s\in[m]$. The second term in RHS of (\ref{eq:hatDeltas_ineq2}) is related to $\sum_{s=1}^m |\Psi_s|$, which is the model complexity of the heterogeneity matrices. Ignoring the logarithmic factor, the second term in RHS of (\ref{eq:hatDeltas_ineq2}) is also sharp w.r.t. the degrees of freedom.

 The bound (\ref{eq:hatDeltas_ineq2}) actually implies the estimators $\hat \bL_s$ are relatively consistent under mild conditions. For notional clarity,  we denote $\bH_s={\rm diag}(\bh_s)$ and $\hat\bH_s={\rm diag}(\hat\bh_s)$ with $\bh_s, \hat\bh_s\in\RR^n$ for all $s\in[m]$. For ease of exposition, we denote $h_s=\min_{i\in[n]} |\bh_s(i)|$ for $\forall s\in[m]$. Denote $\hmin=\min_{s\in[m]} h_s$ and $\hmax=\max_{s\in[m]} h_s$.  The bound (\ref{eq:hatDeltas_ineq2}), together with the fact that $\|\bL_s\|_{\rm F}^2\geq \hmin^4 n^2/r$, yields 
$$
\frac{\sum_{s=1}^m \|\widehat \bL_s-\bL_s\|_{\rm F}^2}{\sum_{s=1}^m \|\bL_s\|_{\rm F}^2}=O\bigg(\frac{r^2}{\hmin^4 nm}\cdot \max_{s\in[m]}\sigma_s^2+\frac{r\sum_{s=1}^m \sigma_s^2|\Psi_s|\log n}{\hmin^4n^2m}\bigg).
$$
The relative error in Frobenius norm diminishes as long as $\hmin^4 nm\gg r^2\max_{s\in[m]}\sigma_s^2+n^{-1}\sum_{s=1}^m \sigma_s^2|\Psi_s|\log n$, which holds trivially if further assuming $\max_{s\in[m]}\sigma_s^2=O(\hmin^4)$. Moreover, the error rate decays as $m$ increases under the upper bound condition on the support cardinality $|\Psi_s|$. 

\begin{rmk} 
Multi-view versions of stochastic block models have been studied in \cite{agterberg2022joint,lei2020consistent,paul2020spectral}, showing that, under suitable conditions, increasing the number of observed networks can improve the accuracy in estimating the shared model parameters. While such an improvement is also achieved by our method and model, as discussed above, a crucial difference in our msLBM model is its accommodation of between view heterogeneity. In fact, the total number of heterogeneity parameters, of order $O(nm)$, increases with the number of views $m$, unlike the models in the aforementioned works, whose complexities typically remain constant irrespective of $m$. The inclusion of these additional heterogeneity parameters in the msLBM enhances the model's robustness, albeit at the cost of more complex estimation procedures and theoretical investigations. 
\end{rmk}

\subsection{Spectral Clustering Consistency}
 As discussed after Theorem~\ref{thm:W-Theta-err}, the low-rank part $\bL_s$ of each view can be consistently recovered under mild conditions. We now investigate the performance of the estimated shared correlation matrix $\hat{\bC} = \hat{\bU}\hat{\bU}^{\top}$. The following theorem shows that, as long as the weights are properly chosen, the consensus graph can be estimated more accurately when more views of data sources are available.

\begin{theorem}\label{thm:hatC-err}
Suppose the conditions of Theorem~\ref{thm:W-Theta-err} hold, $\hmin^4\geq \max_{s}\sigma_s^2\cdot C_0'(\hmax/ \hmin)^4r^2(r+m)$ and $C_1'(\hmax/\hmin)^4\sum_{s=1}^m h_s^{-4}\sigma_s^2r^2|\Psi_s|\log n\leq n$ for  large but absolute constants $C_0', C_1'>0$ depending only on $\kappa_1$, if $\alpha_s\asymp h_s^{-4}$, the following bound holds with probability at least $1-mn^{-2}$,
$$
\frac{\|\hat\bC-\bC\|_{\rm F}^2}{n^2}\leq  C_2\frac{r^2}{nm}\cdot \big(\max_{s\in[m]} \alpha_s \sigma_s^2\big)+\frac{C_3r}{mn^2}\sum_{s=1}^m \alpha_s \sigma_s^2|\Psi_s|\log n,
$$
where $C_2,C_3>0$ are absolute constants depending on $\kappa_1$ only. Meanwhile, we have
$$
\min_{\bO\in\OO_{r\times r}}\|\hat\bU-\bU\bO\|_{\rm F}^2\leq C_4\frac{r^3}{m}\cdot \big(\max_{s\in[m]} \alpha_s \sigma_s^2\big)+\frac{C_5r^2}{mn}\sum_{s=1}^m \alpha_s \sigma_s^2|\Psi_s|\log n,
$$
where $C_4,C_5>0$ are absolute constants depending on $\kappa_1$ only
\end{theorem}
In the case $r, m=O(1)$ and $\hmax/\hmin=O(1)$, the first condition of Theorem~\ref{thm:hatC-err} becomes $\hmin^4\gtrsim \max_{s\in[m]}\sigma_s^2$. This requires that the diagonal entries of $\bW_s$ should be larger than the noise standard deviation. It is a very mild condition in EHR where the observed diagonal entries are often significantly dominating. 

 Recall that $\|\bU\|_{\rm F}\asymp n^{1/2}$. For general $r$ and $m$ satisfying the conditions of Theorem~\ref{thm:hatC-err}, we have $\|\bC\|_{\rm F}\gtrsim n/\sqrt{r}$. 
Theorem~\ref{thm:hatC-err}, together with the conditions of $\hmin$, implies that  
$$
\frac{\|\hat\bC-\bC\|_{\rm F}^2}{\|\bC\|_{\rm F}^2}=O\bigg(\frac{r^3}{nm}\cdot \max_{s\in[m]}\frac{\sigma_s^2}{h_s^4}\bigg),
$$
which holds with probability at least $1-mn^{-2}$. 
Interestingly, it suggests that the relative error decreases as either $n$ or $m$ or both increase. Thus, integrating more data sources can improve the estimation of the correlation matrix $\bC$. We note that the sub-optimal term $r^3$ in the above bound is due to the technical difficulty in bounding the sup-norm error rate $\|\hat\bDelta_s\|_{\ell_\infty}$, which is derived by exploiting the incoherence properties of $\bL_s$'s.

Under similar conditions, we can also get 
$$
\min_{\bO\in\OO_{r\times r}} \frac{\|\hat\bU-\bU\bO\|_{\rm F}^2}{\|\bU\|_{\rm F}^2}=O\bigg(\frac{r^3}{nm}\cdot \max_{s\in[m]}\frac{\sigma_s^2}{h_s^4}\bigg)
$$ 
implying that the factor $\bU$ can be consistently recovered if $nm/r^3\to\infty$.  The rows of $\bU$ provide the information of cluster memberships of vertices. Now we study the clustering error based on its empirical counterpart $\hat\bU$. 

We apply the K-means algorithm on $\widehat\bU$ to get the approximate $(1+\eps)$ solution as in (\ref{eq:app_Kmeans}). Let $\widehat\bZ\in\scrZ_{n,K}$ denote the output membership matrix. In this section, we show that the proposed algorithm in Section~\ref{sec:method} can consistently recover the latent membership matrix under the minimal SNR condition. For two membership matrices $\bZ_1, \bZ_2\in\scrZ_{n,K}$, define the mis-clustering number as 
$$
e_1(\bZ_1,\bZ_2)=\min_{\bP\in\scrP_{K,K}} \frac{\|\bZ_1\bP-\bZ_2\|_{\ell_1}}{2}
$$
where $\scrP_{K,K}$ denotes the set of all $K\times K$ permutation matrices. 

\begin{theorem}\label{thm:cluster}
Suppose the conditions of Theorem~\ref{thm:hatC-err} holds and $m\nmin^2\delta_{\bOmega}^2\geq C_0r(r+m)n\max_{s}(\alpha_s \sigma_s^2)+C_1r\sum_{s=1}^m \alpha_s \sigma_s^2|\Psi_s|\log n$ for large constants $C_0, C_1$ depending on $\kappa_1$ only, the following bound holds with probability at least $1-mn^{-2}$,
$$
\frac{e_1(\hat\bZ,\bZ)}{n}\leq \frac{4(2+\eps)^2}{\nmin \delta_{\bOmega}^2}\cdot \bigg[C_2\frac{r^2}{m}\cdot \big(\max_{s\in[m]} \alpha_s \sigma_s^2\big)+\frac{C_3r}{mn}\sum_{s=1}^m \alpha_s \sigma_s^2|\Psi_s|\log n\bigg],
$$
where $C_2,C_3>0$ depends only on $\kappa_1$. 
\end{theorem}
In the case that $\nmin\asymp nK^{-1}$ and $\eps\leq 1$,  together with the condition of $\hmin$ in Theorem~\ref{thm:hatC-err}, the mis-clustering relative error becomes
\begin{align}\label{eq:e1hatZ-err-1}
\frac{e_1(\hat\bZ,\bZ)}{n}=O\Big(\frac{K}{mn\delta_{\bOmega}^2}\Big),
\end{align}
which converges to $0$ if $mn\delta_{\bOmega}^2/K\to\infty$ as $nm\to\infty$. The number of clusters $K$ in the motivating EHR application is large, e.g., around $O(n^{1/2})$ in as seen in Section~\ref{sec:realdata}.  However, if $K$ is small such that $K=O(m)$, and moreover if $\delta_{\bOmega}\gtrsim 1$, we get $e_1(\hat\bZ,\bZ)=O(Km^{-1})$ implying that $e_1(\hat\bZ,\bZ)=O(1)$. This is interesting since it suggests that our algorithm can exactly recover nearly all of the vertices' membership.

\subsection{Consensus Graph Estimation} 
We next establish the error of 
\begin{equation}\label{eq:hatOmega}
\widehat\bOmega=(\widehat\bZ^{\top}\widehat\bZ)^{-1}\widehat\bZ^{\top} (\widehat\bU \widehat\bU^{\top})\widehat\bZ(\widehat\bZ^{\top}\widehat\bZ)^{-1}
\end{equation}

\begin{theorem}\label{thm:Omega}
Suppose that the conditions of Theorem~\ref{thm:cluster} hold such that $e_1(\hat\bZ,\bZ)\leq \nmin/2$, there exist constants $C_3,C_4>0$ depending only on $\kappa_1$ such that
\begin{align*}
\|\hat\bOmega-\hat\bP\bOmega\hat\bP^{\top}\|_{\rm F}^2\leq (2+\eps)^2\frac{\|\bOmega\|^2}{\delta_{\bOmega}^2}\cdot \frac{\nmax^{3}n}{\nmin^{5}}\cdot\bigg[C_3\frac{r^2}{m}\cdot \big(\max_{s\in[m]} \alpha_s \sigma_s^2\big)+\frac{C_4r}{mn}\sum_{s=1}^m \alpha_s \sigma_s^2|\Psi_s|\log n\bigg]
\end{align*}
where $\widehat\bP$ is the $K\times K$ permutation matrix realizing $\min_{\bP\in\scrP_{K,K}}\|\bZ-\widehat\bZ \bP\|_{\ell_1}$.
\end{theorem}
The condition $e_1(\hat\bZ,\bZ)\leq \nmin/2$ is mild. Indeed, by (\ref{eq:e1hatZ-err-1}), this condition holds as long as $nm\delta_{\bOmega}^2\gtrsim K^2$ in the case $\nmin\asymp\nmax\asymp nK^{-1}$. If the $\delta_{\bOmega}$ is bounded away from $0$, then the number of clusters is allowed to grow as fast as $(nm)^{1/2}$. 
Together with the condition on $\hmin$ in Theorem~\ref{thm:hatC-err}, the bound in Theorem~\ref{thm:Omega} implies that 
$$
\frac{\|\hat\bOmega-\hat\bP\bOmega\hat\bP\|_{\rm F}^2}{\|\bOmega\|^2}=O\Big(\frac{K^2}{mn\delta_{\bOmega}^2}\Big),
$$
which converges to zero as long as  $mn\delta_{\bOmega}^2\to\infty$ as $mn\to\infty$.

\section{Simulations}\label{sec:simulations}

In this section, we present simulation results to evaluate the finite sample performance of the proposed msLBM estimator obtained through Algorithm \ref{algo:exact} and compare it to existing methods. Throughout we set $m =3, n=500, r=25$. For simplicity, we considered a balanced underlying clustering structure such that $n_k \approx n/K, 1\leq k\leq K$ under a range of $K=25,50,75,100$. For each configuration setting, we summarize results based on the average from 50 independent experiments. 

To mimic a real-world sparse network, we first generate a sparse matrix $\bA = [A(i, j)]_{K \times r}$ with normalized rows and then set $\bOmega = \bA \bA^\top$. Specifically, we generate {$A_{i,j} \stackrel{i.i.d.}{\sim} (1-\pi_0)\cdot {\bf 1}_{ \{x=0 \}}+\pi_0 \cdot \mbox{U}(0,1)$ for $i \in [K]$ and $j \in [r]$ where $\pi_0 = 0.2$ to be comparable to what we observed in the real data analysis. Then we normalize the rows of $\bA$ to make all of its rows have unit $\ell_2$ norm. We then fix $\bOmega$ for all of the repetitions. 

We consider two settings for generating $\bH\subbullet$ and $\bTheta\subbullet$:  setting (1) representing a heterogeneous view  $\bH_s$ and $\bTheta\subbullet \ne \bzero$ and setting (2) representing a more homogeneous scenario with $\bH_s = \lambda \sqrt{s} \bI_{n}$ and $\bTheta_s = \bzero$.  

In  setting (1), we generate $\{\bTheta_s\}_{s\in [m]}$ by sampling its entries independently from the distribution $(1-\pi){\bf 1}_{ \{x=0 \}} + \pi\cdot N(0,\tau^2)$ with $\pi=0.05$ and $\tau = 5$ here. 
To show that our algorithm is useful for a wide range of $\bH_s$, we generate the diagonal entries of $\bH_s$ from $\mbox{Uniform}[0,d_s]$ where $(d_1,d_2,d_3) = \lambda (1,\sqrt{2},\sqrt{3})$ and $\lambda$ is chosen between $1$ and $2$ to represent varying signal strengths. In the homogeneous setting (2),  the three views share a common eigenspace but have different relative signal strengths. We let $\lambda$ vary from  $0.25$ to $1$ to reflect different levels of signal strengths. Finally, given $\bOmega$, $\bH\subbullet$ and $\bTheta\subbullet$, we generate the sparse error matrix $\bE_s$ by sampling its entries independently from   $0.5 {\bf 1}_{ \{x=0 \}} + 0.5 \cdot N(0,\tau^2)$. For setting (1),  $\sigma_s=0.1$ for $s \in [m]$, and for setting (2), $(\sigma_1,\sigma_2,\sigma_3) = (0.3,0.2,0.1)$.

To evaluate the performance of our proposed and benchmark methods, we consider the ability of the methods to recover $\bZ$, $\bOmega$, $\bTheta_s$, and the eigenspace of $\bC$,  respectively.  More specifically, the mis-clustering error  (MCE) for $\bZ$ is defined as
$$\MCE(\widehat \bZ, \bZ) = \min_{\bP\in\scrP_{K,K}} {1 \over 2n}\|\widehat{\bZ}\bP-\bZ\|_{\ell_1},$$
where $\scrP_{K,K}$ denotes the set of all $K\times K$ permutation matrices. We consider the $\ell_2$ loss for $\bOmega$ and the $\ell_0$ loss for the sparse matrices  $\{\bTheta_s\}_{s\in [m]}$, defined as
$$ \sfL_2(\widehat\bS, \bS) = \frac{\|\widehat\bS - \bS\|_{\rm F}^2}{\|\bS\|_{\rm F}^2}
\quad\mbox{and}\quad \sfL_0(\widehat\bS, \bS)= \frac{\|{\rm supp}(\widehat\bS) - {\rm supp}(\bS) \|_{\ell_1}}{K^2},$$
for any matrix $\widehat \bS$ and $\bS$. In addition, given a pair of matrices with orthonormal columns $\bV$ and $\hat{\bV}$ $\in \scrO_{n,r}$, we measure the distance between their invariant subspaces via the spectral norm of the difference between the projections, given by $\|\hat{\bV}\hat{\bV}^\top  - \bV\bV^\top \|$.

Since no existing methods consider the same model as ours, we compare to some relevant methods that can be used to identify the group structure $\bZ$. Specifically, we compare to (i) the sum of adjacency matrices (SAM) approach  \citep{bhattacharyya2018spectral} that estimates the common eigenvectors and the group structures $\bZ$ of multi-view networks; and (ii) the MASE approach on multiple low-rank networks with common principal subspaces \citep{arroyo2021inference}. After performing SVD on the sum of the adjacency matrices to obtain $\hat{\bU}$, the SAM algorithm identifies $\bZ$ by either performing a K-means clustering algorithm on the row vectors of $\hat{\bU}$ (SAM-mean) or K-median clustering algorithm on the normalized row vectors of $\hat{\bU}$ (SAM-median).  We only keep $\hat\bU$ to be rank $r$ since the full rank version performs poorly in our settings. We include both MASE and the scaled version of MASE (MASE-scaled) as proposed in \cite{arroyo2021inference}.  In addition, we compare the ASALM algorithm applied to a single source in recovering $\bOmega$ and $\bTheta_s$. To be specific, the estimated sparse matrix by ASALM is exactly the estimator of $\bTheta_s$ in each view. We then input the estimated low-rank matrix by ASALM to Algorithm \ref{algo:hatUs} to obtain the estimator of $\widehat \bU$ for each view. Then we use the $\widehat \bU$ to estimate $\widehat \bZ$ by \eqref{eq:app_Kmeans} and $\widehat \bOmega$ by \eqref{eq: Omegahat}.

For setting (1), we only compare the clustering performance of SAM-mean, SAM-median, MASE, and MASE-scaled to our algorithm since under setting (1), these views do not share a common eigenspace and these competing methods assume a common eigenspace across different views.
In addition, we compare the $\ell_2$ loss for $\bOmega$ and the $\ell_0$ loss for the sparse matrices  $\{\bTheta_s\}_{s\in [m]}$ of msLBM and ASALM, averaged over the three views. For setting (2), we compare the clustering performance of the four methods mentioned above as well as the eigenspace error of the leading $r$ eigenvectors from SAM, MASE, and MASE-scaled.

We choose the tuning parameters using the procedure detailed in Section \ref{tuning} by randomly sampling $500$ entries from $\bC$ plus normal noise $N(0,0.01^2)$. This procedure mimics our real data example where we have several sets of human annotated similarity and relatedness of vertex pairs. See Section \ref{sec:realdata} for details. For the choice of the number of groups $K$, to validate the tuning strategy in Section \ref{tuning}, we focus on setting (1) with $\lambda = 1.25$ and $K = 100$, which is the most difficult task due to the lower signal to noise ratio and the large $K$.  To be specific, we randomly sample $100$ positive pairs ($100$ pairs of vertices within groups) and $1000$ negative pairs with a correlation larger than $0.5$ ($1000$ pairs of vertices between groups). Then for each $K$, we can get $\widehat \bZ$ from \eqref{eq:app_Kmeans} and finally choose the $K$ achieving the optimal composite score. The procedure is repeated $50$ times, and the average optimal $K$ is $100.28$ with a standard deviation of $7.83$. Due to the effectiveness of the method, we decide to treat $K$ as known during the simulation.

We first compare the MCE of msLBM, SAM-mean, SAM-median, MASE and MASE-scaled. The result is shown in Figure \ref{fig:SS3}.  In general, the SAM-mean, MASE, and MASE-scaled perform the worst while msLBM wins across all scenarios. In addition, SAM-mean performs better than the other three comparable methods.  It is reasonable that the SAM-mean is designed for degree corrected model which is more suitable for the current case. In addition, SAM and MASE require a common eigenspace across all views but our model does not necessarily satisfy the assumption. It is apparent that msLBM significantly outperforms other methods with even more advantages as $K$ becomes larger.

\begin{figure}[htbp!]
	\begin{center}
		\includegraphics[width=0.98\textwidth]{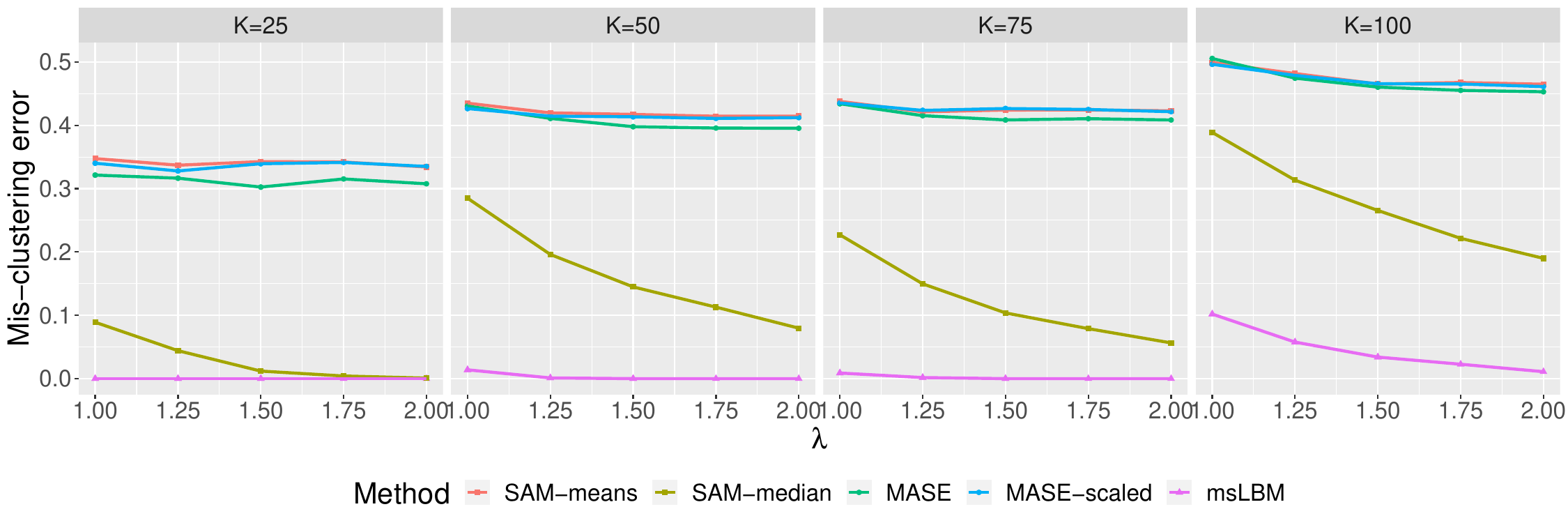}
		\caption{Comparison on clustering performance under setting (1).}
		\label{fig:SS3}
	\end{center}
\end{figure} 

Then, we evaluate the performance of $\widehat\bOmega$ and $\{\widehat\bTheta_s\}_{s\in [m]}$ estimated from msLBM and ASALM. The result is shown in Figure \ref{fig:l2omega}. Again, the msLBM performs much better than ASALM. This is intuitive given that msLBM utilizes the information of all views while ASALM does not. 

\begin{figure}[htbp]
	\begin{center}
		\includegraphics[width=0.98\textwidth]{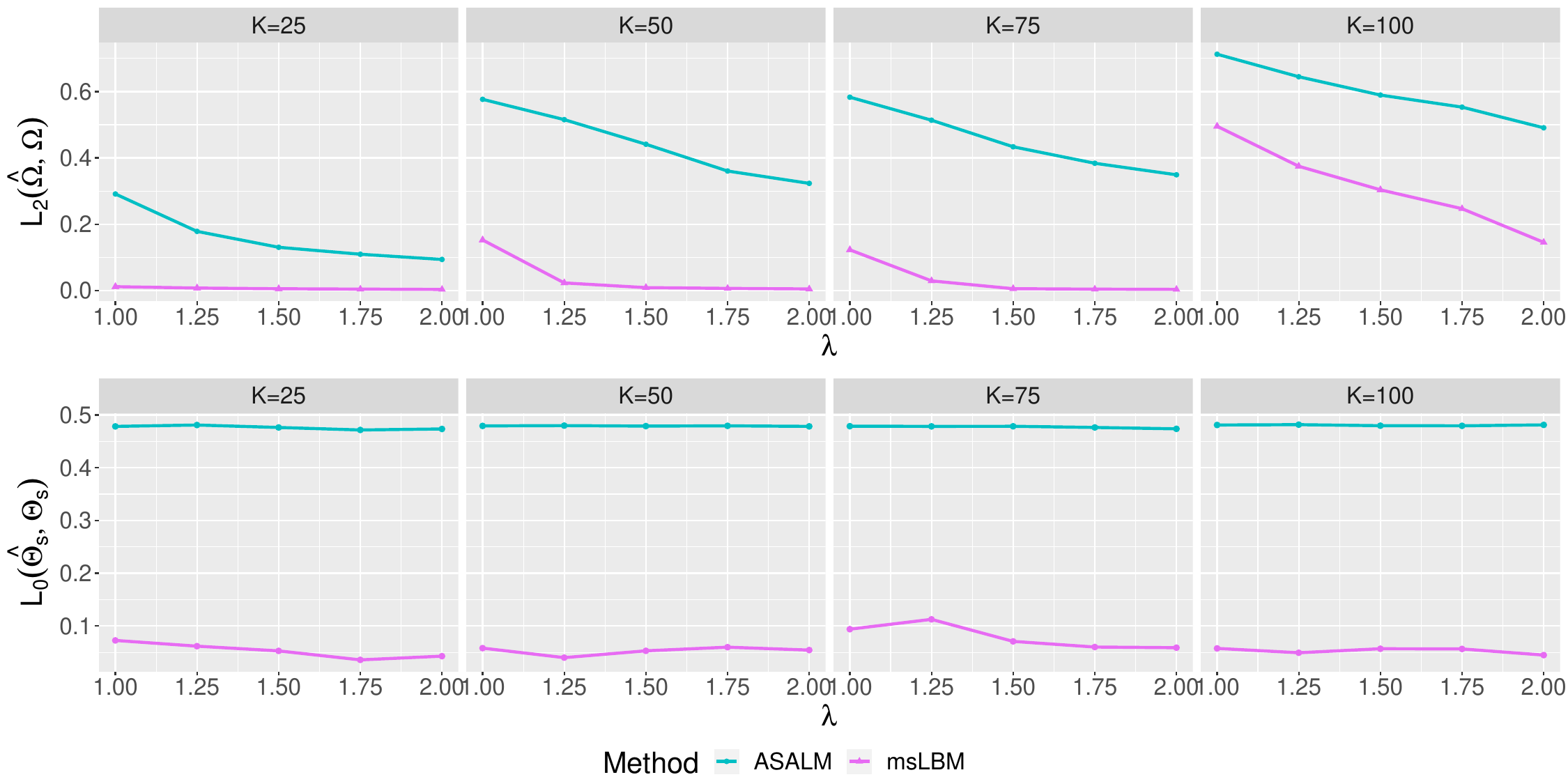}
		\caption{Comparison on recovering $\bOmega$ (top panel) and $\bTheta_s$ (bottom panel) under setting (1).}  \label{fig:l2omega}
	\end{center}
\end{figure}

Finally, we present the mis-clustering error and the eigenspace of these methods under setting $2$ in Figure \ref{fig:Z2}, when the three views share a common eigenspace. Even though the current models also satisfy the assumption of the completing methods SAM and MASE, msLBM still performs much better than them because it can fully exploit the relative signal and noise strength of each view. On the contrary, SAM can not achieve an optimal weighted combination of all views by simple average. MASE-scaled can use the relative signal strength of each view by using the eigenvalues of each view. However, it is not robust when the signal strength is small compared to the noise level. So we observe that when $\lambda$ is small, it performs worst. But when $\lambda$ increases, it performs better than MASE. MASE performs worst when $\lambda$ is large because it is unweighted.

\begin{figure}[htbp]
	\begin{center}
		\includegraphics[width=0.98\textwidth]{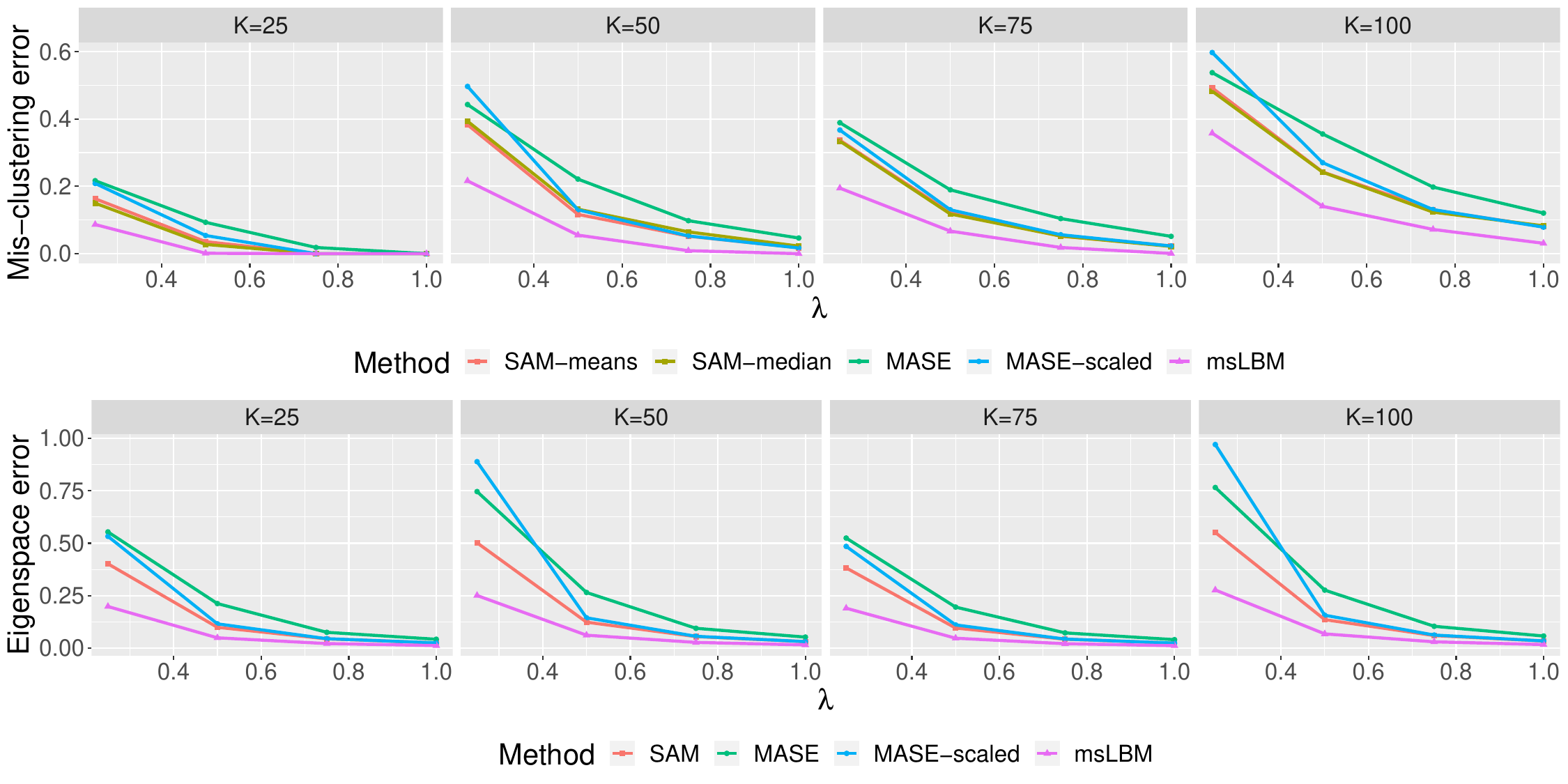}
		\caption{Comparison on recovering $\bZ$ (top panel) and the eigenspace of $\bC$ (bottom panel) under setting (2).} \label{fig:Z2} 
	\end{center}
\end{figure}

\section{Applications to learning clinical knowledge graph}
\label{sec:realdata}

We next apply the proposed msLBM method to learn both embeddings for medical concepts and a consensus clinical knowledge graph by synthesizing a few sources of medical text data.

\subsection{Data summary}
\def \PMI {\rm PMI}
\def \SPPMI {\rm SPPMI}
\def \CUI {\rm CUI}

The input data ensemble consists of three similarity matrices of $n=7,217$ clinical concepts, independently derived from three heterogeneous data sources: (i) 10 million clinical narrative notes of 62K patients at Partners Healthcare System (PHS); (ii) 20 million clinical notes from a Stanford hospital \citep{finlayson2014building}; (iii) clinical notes from the MIMIC-III (Medical Information Mart for Intensive Care) database \citep{Johnson_2016}. The clinical concepts were extracted from textual data via natural language processing by mapping clinical terms to Concept Unique Identifiers ($\CUI$s) from UMLS. Heterogeneity inherently plays a role due to the different natures of data sources. For instance, MIMIC-III data only involves intensive care unit (ICU) patients. The frequency of disease conditions seen in ICU settings differs from those in the other two views. These datasets also have very different sample sizes. As a result, the marginal frequencies $\bH_s$ and the noise level $\sigma_s$ can vary greatly across views. The management of disease conditions, such as the treatment used for a given condition, could also differ across health systems and hospital settings (e.g., ICU vs. outpatient), necessitating $\bTheta_s$ to account for this type of population-level bias. 

For each source $s$, we construct $\bW_s$ as the shifted-positive pointwise mutual information (SPPMI) matrix for all $n$ concepts. 
For a pair of concepts $x$ and $y$, the pointwise mutual information ($\PMI$) is a well-known information-theoretic association measure between $x$ and $y$, defined as $$\PMI(x,y) =  \log \frac{P(x,y)}{P(x)P(y)}$$
where $P(x,y)$ is the probability of $x$ and $y$ co-occurring and $P(x), P(y)$ are respective marginal occurrence probabilities. Using $\PMI$ as a measure of association in NLP was introduced by \cite{church1990word} and has been widely adopted for word similarity tasks \citep[e.g.]{dagan1994similarity, turney2010frequency}. However, the empirical $\PMI$ matrix is not computationally or statistically feasible to use since the PMI estimate would be $-\infty$ for a pair that never co-occurs and the matrix is also dense. The SPPMI matrix with $\SPPMI(x, y) = \max \{ \widehat{\PMI}(x,y), 0 \}$ is a sparse and consistent alternative estimate of PMI widely used in the NLP literature \citep{levy2014neural}.  

Previous studies \citep[e.g.]{levy2014neural,arora2016latent} have noted that the SPPMI matrix often exhibits low-rank characteristics, which can be approximated by the inner product of word embeddings under the dynamic log-linear topic model proposed by \cite{arora2016latent}. Our model assumption stems from the use of cosine similarities (instead of inner products) of word embeddings. Cosine similarities between two features should remain consistent across different views or sources. However, the norms of the embeddings, represented by $\bH_s$ can vary to reflect the heterogeneity in the marginal occurrence probabilities of these concepts in the generative model of \cite{arora2016latent}. Taking into account that various healthcare systems might display distinct frequencies of the same concept and have unique patterns inherent to each system, it is logical to consider different $\bH_s$ for different views. Additionally, incorporating the sparse matrix $\bTheta_s$ is beneficial to accommodate the view-specific patterns that the low-rank component does not explain and may differ across sites, thereby enhancing the robustness of the model.

We demonstrate below how Algorithm~\ref{algo:exact} can be used to optimally combine information from the SPPMI matrices from the three data sources to both improve the estimation of embeddings and construct a sparse knowledge graph about coronary artery disease network. We aim to group the $7,217$ CUIs into $K$ subgroups such that CUIs within the same subgroups are considered synonyms. We anticipate $K$ to be large in this particular application since CUIs were manually curated to represent distinct clinical concepts. 

\subsection{Clinical concept Embedding and Network}
\label{sec:6.2}
With the three $\SPPMI$ matrices as input, our Algorithm \ref{algo:exact} will output the clinical concept network $\hat{\bC}$. $\hat{\bC}$ has the expression $\hat{\bC} = \hat{\bU} \hat{\bU}^\top$ where $\hat{\bU} \in \RR^{n \times r}$ is the spherical embedding of the $n$ clinical concepts. We will present that $\hat{\bC}$ (or equivalently $\hat{\bU}$) have better quality than using single source only or other comparable methods. 

At first, we can compare the result of Algorithm \ref{algo:exact} to the clinical concept learned from a single source. Given $\bW_s$, we can apply Algorithm \ref{ReC} to it to get the clinical concept network. In addition, we can compare our method to SAM and MASE. For SAM, we get the $\hat{\bV}_{\textsf{SAM}} \in \RR^{n \times r}$ consisting of the leading $r$ eigenvectors of the sum of the three $\SPPMI$ matrices. Then we normalize its row to get $\tilde{\bV}_{\textsf{SAM}}$, the spherical embedding and the clinical concept network $\hat{\bC}_{\textsf{SAM}} = \tilde{\bV}_{\textsf{SAM}}\tilde{\bV}_{\textsf{SAM}}^\top$. For MASE (or MASE-scaled), when we get $\hat{\bV}_{\textsf{MASE}}$, we also normalize its rows to get $\tilde{\bV}_{\textsf{MASE}}$ and get $\hat{\bC}_{\textsf{MASE}} = \tilde{\bV}_{\textsf{MASE}}\tilde{\bV}_{\textsf{MASE}}^\top$. To fully highlight msLBM's ability to exploit the relative signal and noise strengths of each view, we also include the collapsing method (Collapse). This method estimates each of the three probabilities in PMI by utilizing all views and then applies Algorithm \ref{algo:hatUs} to the estimated SPPMI matrix.

To assess the quality of the estimated concept networks, we refer to two sets of human evaluations released by \cite{pakhomov2010semantic}, which focus on the semantic similarity and relatedness between clinical concepts. Semantic similarity measures the extent of semantic overlap between concepts based on psycholinguistic definitions (for example, 'arthritis' versus 'joint pain'), while relatedness pertains to the likelihood of one concept evoking thoughts of another (like 'diabetes' versus 'metformin'). We evaluate the alignment between human annotations and the estimated clinical concept networks by calculating the Spearman rank correlation between human assessments of similarity/relatedness among CUI pairs and the correlations between CUI pairs in the estimated networks. These human annotations provide a valuable benchmark for assessing the quality of our generated network.

Furthermore, the effectiveness of the concept networks can be gauged through their ability to identify known relational pairs. Specifically, we utilize five sets of CUI relation pairs extracted from a medical database, which include \textbf{May Cause} (MayCause), \textbf{May Be Caused By} (Causedby), \textbf{Differential Diagnosis} (Ddx), \textbf{Belong(s) to the Category of} (Bco), and \textbf{May Treat} (MayTreat). Following the methodology of \cite{beam2018clinical}, we report the Area Under the Curve (AUC) and the true positive rate (TPR), setting the false positive rate (FPR) at $1\%$, $5\%$, and $10\%$,   respectively, to measure performance.

To select the appropriate rank $r$, we examine the eigenvalue decay of $\{\bW_s\}_{s=1}^m$. This technique is commonly employed for identifying the rank of low-rank matrices, as seen in various applications such as principal component analysis \citep{jolliffe2005principal}, word embedding \citep{hong2021clinical}, and network analysis \citep{arroyo2021inference}. We assess the eigenvalue decay of $\{\bW_s\}_{s=1}^m$, opting for the rank $r$ that results in the cumulative eigenvalue percentage exceeding $95\%$ for all matrices. This criterion leads us to choose $r = 250$. Consequently, we apply $r = 250$ across all methods in our analysis. For fine-tuning the remaining parameters, we use the strategy in Section \ref{tuning} using the relatedness set as the noisy labels. 

In addition, we perform a sensitivity analysis by experimenting with various values of $r$ in these methods and assessing their performance. The outcomes of this analysis are depicted in Figure \ref{fig:sen}. This figure indicates that our method maintains a similar level of performance across a broad range of $r$ values and performs adequately when $r \geq 150$. Furthermore, our method consistently outperforms all other methods for all $r$'s. Therefore, in subsequent analyses, we will primarily focus on results obtained with $r=250$.

\begin{figure}[htbp]
	\begin{center}
\includegraphics[width=0.98\textwidth]{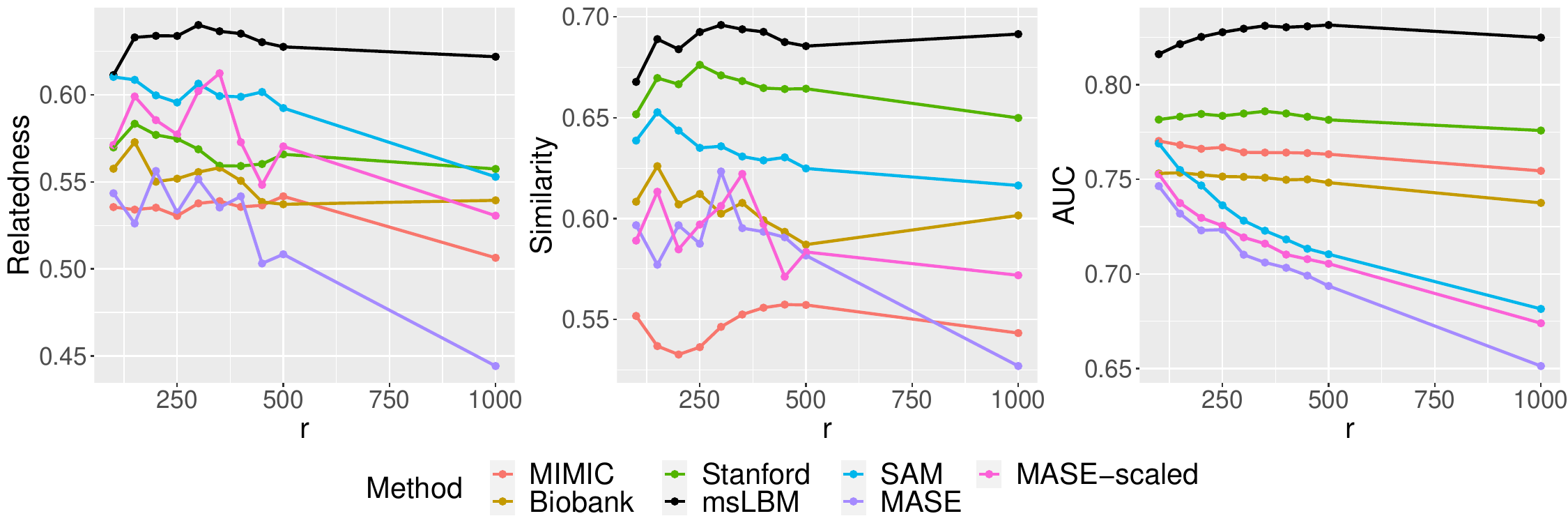}
		\caption{
The Spearman rank correlation is depicted for the relatedness scores (the left panel) and the similarity scores (the middle panel). Additionally, the average AUC is presented for the five types of relation pairs (the right panel).}
		\label{fig:sen}
	\end{center}
\end{figure}

Table \ref{tb:correlation} summarizes the Spearman rank correlation using each single data source and the output given by Algorithm~\ref{algo:exact} with $r=250$. Tables \ref{tab:AUC}, \ref{tab:fpr005}, and \ref{tab:fpr01} show the AUC and TPR given FPR $ =0.05$ and $0.1$ of relation detection, respectively. Synthesizing information from three data sources via  Algorithm~\ref{algo:exact} yields higher quality embeddings compared to a single source concept network as evidenced by the performance of all of these tasks. In addition, our method is better than SAM, MASE, and MASE-scaled. The three methods are even worse than a single source in some cases. A possible reason may be that they can not impose suitable weights on different sources. The collapsing method is similar to SAM in the sense that they try to estimate the common network by aggregating the data from all views. However, since it cannot fully exploit the relative signal and noise strength of each view, its performance is not as good as msLBM. The collapsing method is similar to SAM in that it tries to estimate the common network by aggregating data from all views. However, it fails to fully exploit the relative signal and noise strengths of each view, resulting in inferior performance compared to msLBM.

\begin{table}[htbp]
\centering
\begin{tabular}[t]{l|r|r}
\hline 
Method (Data Source)  & Relatedness & Similarity\\
\hline \hline
MIMIC & 0.533 & 0.547\\
Biobank & 0.539 & 0.605\\
Stanford & 0.567 & 0.660\\
msLBM & \textbf{0.635} & \textbf{0.686}\\
SAM & 0.588 & 0.640\\
MASE & 0.577 & 0.590\\
MASE-scaled & 0.517 & 0.577\\
Collapse & 0.601 & 0.662 \\
\hline
\end{tabular}
\caption{Spearman rank correlation under $r=250$.}
\label{tb:correlation}
\end{table}

\begin{table}[]
\centering
\begin{tabular}[t]{l|r|r|r|r|r}
\hline
Method (Data Source)  & MayCause & Causedby & Ddx & Bco & MayTreat\\
\hline \hline
MIMIC & 0.752 & 0.782 & 0.789 & 0.720 & 0.777\\
Biobank & 0.731 & 0.764 & 0.780 & 0.698 & 0.787\\
Stanford & 0.757 & 0.786 & 0.818 & 0.741 & 0.819\\
msLBM & \textbf{0.802} & \textbf{0.831} & \textbf{0.864} &\textbf{0.804} & \textbf{0.850}\\
SAM & 0.722 & 0.763 & 0.770 & 0.650 & 0.761\\
MASE & 0.709 & 0.750 & 0.773 & 0.637 & 0.752\\
MASE-scaled & 0.700 & 0.739 & 0.762 & 0.647 & 0.751\\
Collapse &  0.770 & 0.806 & 0.820 & 0.795 & 0.821\\
\hline
\end{tabular}
\caption{\label{tab:AUC}AUC of Clinical Relation Detection}
\end{table}

\begin{table}[]
\centering
\begin{tabular}[t]{l|r|r|r|r|r}
\hline  \hline
Method (Data Source)  & MayCause & Causedby & Ddx & Bco & MayTreat\\
\hline \hline
MIMIC & 0.339 & 0.398 & 0.419 & 0.238 & 0.392\\
Biobank & 0.332 & 0.403 & 0.399 & 0.238 & 0.456\\
Stanford & 0.319 & 0.385 & 0.427 & 0.233 & 0.465\\
msLBM & \textbf{0.429} & \textbf{0.495} & \textbf{0.562} & \textbf{0.374} & \textbf{0.559}\\
SAM & 0.371 & 0.439 & 0.474 & 0.239 & 0.472\\
MASE & 0.333 & 0.398 & 0.450 & 0.205 & 0.432\\
MASE-scaled & 0.336 & 0.396 & 0.448 & 0.221 & 0.444\\
Collapse & 0.364 & 0.434 & 0.436 & 0.292 & 0.477 \\
\hline
\end{tabular}
\caption{\label{tab:fpr005}TPR of Clinical Relation Detection with fixed FPR=0.05}
\end{table}

\begin{table}[]
\centering
\begin{tabular}[t]{l|r|r|r|r|r}
\hline \hline
Method (Data Source)  & MayCause & Causedby & Ddx & Bco & MayTreat\\
\hline \hline
MIMIC & 0.451 & 0.508 & 0.523 & 0.359 & 0.500\\
Biobank & 0.443 & 0.511 & 0.515 & 0.356 & 0.558\\
Stanford & 0.458 & 0.529 & 0.565 & 0.420 & 0.577\\
msLBM & \textbf{0.540} & \textbf{0.604} & \textbf{0.672} & \textbf{0.516} & \textbf{0.661}\\
SAM & 0.451 & 0.515 & 0.542 & 0.330 & 0.542\\
MASE & 0.417 & 0.480 & 0.538 & 0.275 & 0.514\\
MASE-scaled & 0.409 & 0.473 & 0.531 & 0.309 & 0.514\\
Collapse & 0.490 & 0.554 & 0.568 & 0.480 & 0.595\\
\hline
\end{tabular}
\caption{\label{tab:fpr01}TPR of Clinical Relation Detection with fixed FPR=0.1}
\end{table}

\subsection{Coronary artery disease network}
We next construct a disease network for Coronary Artery Disease (CAD), a leading cause of death involving multiple progression states. We set $K=1000$ and Figure \ref{fig: decay} suggests the disease network related to CAD is sparse since the magnitude of $\widehat\bOmega$ associated with the CAD CUI decays very fast. To further visualize the network, we focused on a subset of $371$ CUIs that can have been previously identified as potentially related to CAD from $5$ publicly available knowledge sources -- including Mayo, Medline, Medscape, Merck Manuals, and Wikipedia -- as in \cite{yu2016surrogate}.  Algorithm~\ref{algo:exact} grouped these CUIs into $86$ groups. We present in Figure~\ref{fig: CAD DST } the CUIs groups that are most important for CAD as measured by the magnitude of $\widehat\Omega_{kl}$ and in Figure~\ref{fig: CAD cloud } the CUIs included in each of the CUI groups. Our method can yield a very insightful network that unearths the progression states: Hyperlipidemia $\gg$ Atherosclerosis $\gg$ Angina $\gg$ Myocardial Infarction (MI) $\gg$ Congestive Heart Failure (CHF). Associated symptoms such as chest pain are also identified. In addition, our network successfully identifies medications important for CAD including Nitrate: for Angina and Myocardial Infarction; Beta-Blocker for MI and CHF; Anti-platelet for CAD, Angina, and MI.  In addition to recovering the disease network, our method also successfully grouped near identical concepts into meaningful concept groups as shown in Figure~\ref{fig: CAD cloud }. For example, the CAD group consists of multiple synonymous concepts including ``C0010068" for coronary heart disease, ``C0010054" for coronary arteriosclerosis, ``C0151744" for myocardial ischemia as well as ``C0264694" for chronic myocardial ischemia. All these concepts are frequently used in clinical notes to describe CAD. We observe that CUIs indicative of chest pain have been split into two groups named ``Chest Pain'' and ``Chest Discomfort'' respectively by our method.  While it might be ideal from a clinical perspective to merge them into a single chest pain concept group, such little defect is acceptable due to data quality and more importantly its data-driven nature that will not affect the quality of the overall learned network. 

\begin{figure}[htbp]
\begin{center}
\includegraphics[width=0.65\textwidth]{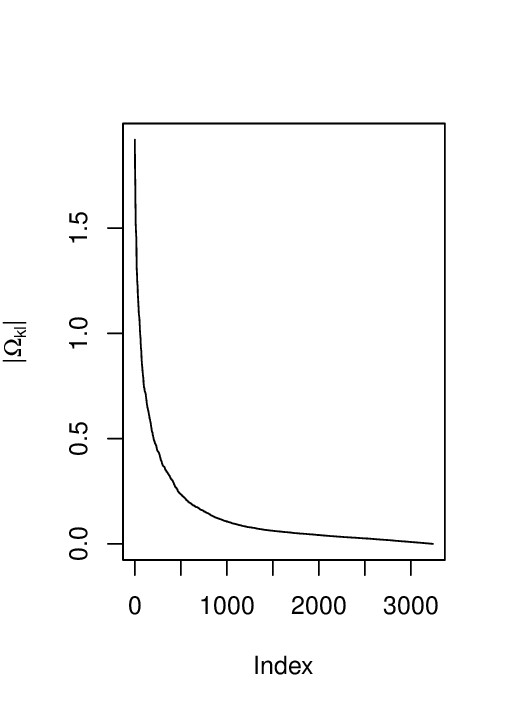}
\caption{$|\Omega_{kl}|$ decay.}\label{fig: decay}
\end{center}
\end{figure}

\begin{figure}[htbp]
\begin{center}
\includegraphics[width=0.95\textwidth]{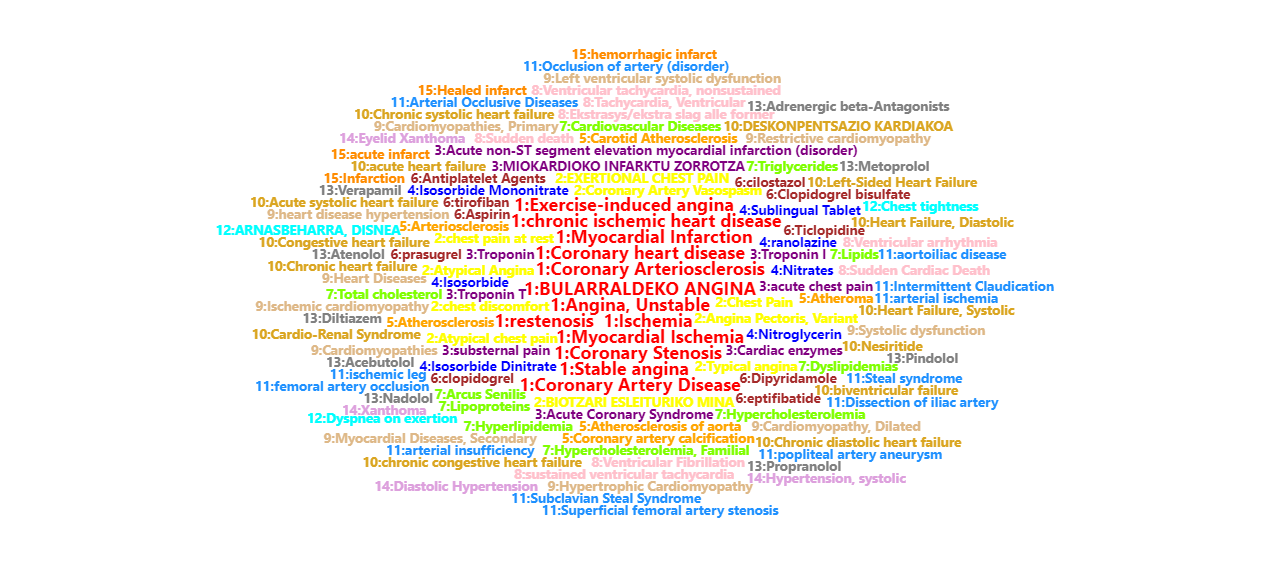}
\caption{CAD Disease network.}\label{fig: CAD DST }
\end{center}
\end{figure}

\begin{figure}[htbp]
\begin{center}
\includegraphics[width=0.95\textwidth]{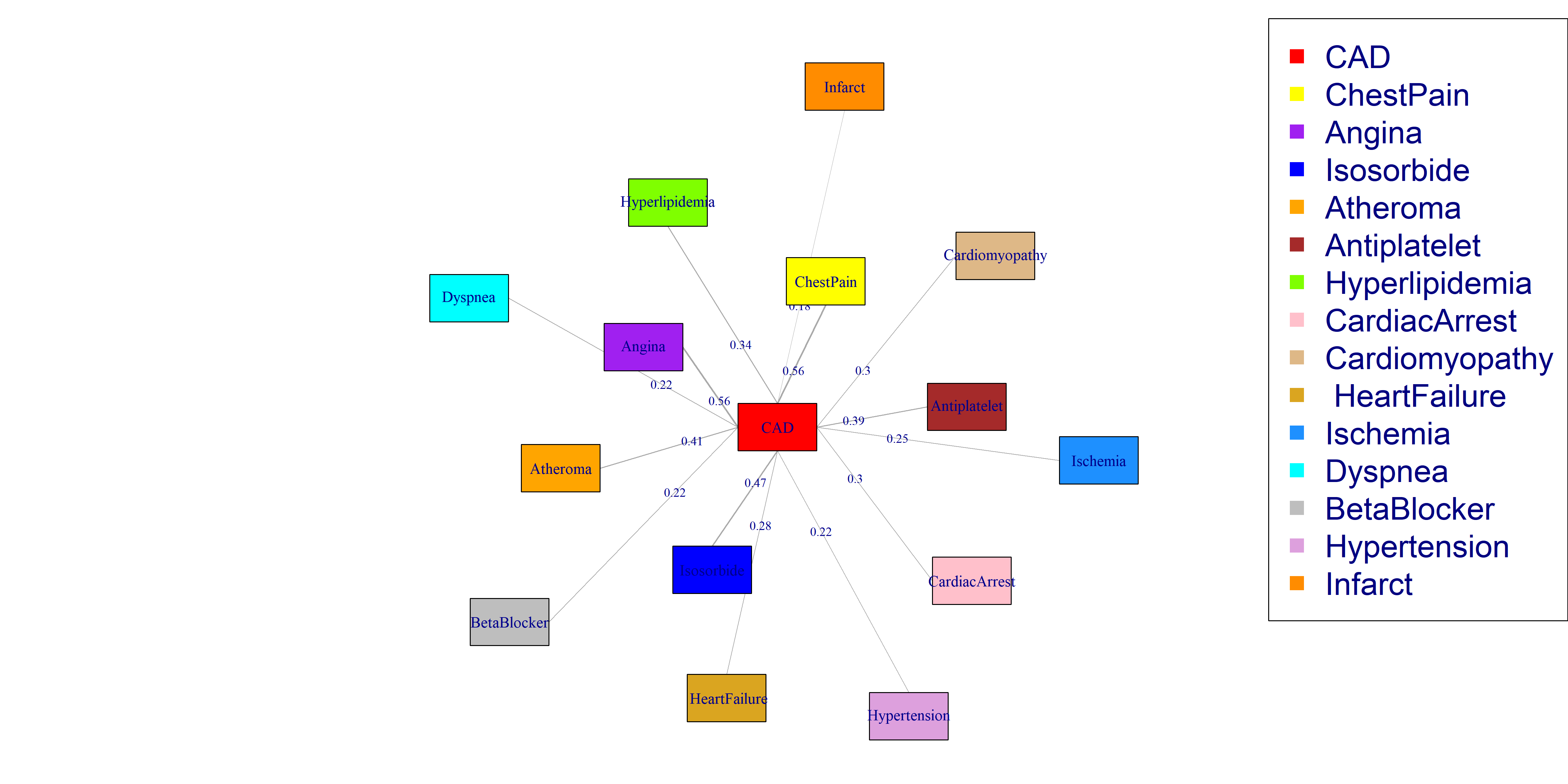}
\caption{Consensus CAD network.} \label{fig: CAD cloud }
\end{center}
\end{figure}

\section{Discussion}\label{sec:discuss}

In this paper, we proposed an msLBM model to synthesize information to learn a consensus graph from multiple sources. Under the msLBM, we developed an alternating minimization algorithm to estimate the unknown parameters associated with the graph and provided convergence properties for the algorithm.  Our model, methodologies, and theories are established under the assumption that $\bC = \bU\bU^{\top}$ is positive semi-definite. This assumption is made for technical convenience. However, we can easily adapt our model, methodologies, and theories to accommodate an asymmetric shared correlation structure. If $\bC$ is asymmetric, we can estimate its balanced factorization $\bC = \bU\bV^{\top}$, where $\bU^{\top}\bU = \bV^{\top}\bV$. The estimating procedures and algorithms can be easily adapted, and the theoretical guarantees will be almost the same as in the positive semi-definite case.

\section{Acknowledgement}
Dong Xia's research was partially supported by Hong Kong RGC Grant GRF 16302020. Doudou Zhou's research was partially supported by NUS Start-up
Grant A-0009985-00-00.

   \bibliographystyle{chicago}
        \bibliography{refer}

\end{document}